\documentclass[11pt]{report}
\usepackage{graphicx}
\usepackage{subcaption}
\usepackage{amsbsy,amsmath,amsfonts,amssymb,graphicx}
\usepackage{algorithm}
\usepackage{algpseudocode}
\usepackage{url}
\usepackage{geometry}
\usepackage[hidelinks]{hyperref}

\usepackage[numbers,sort&compress]{natbib}
\usepackage{booktabs}

\begin{document}

\title{\Huge{Wildfire Autonomous Response and Prediction Using Cellular Automata (WARP-CA)}}

\author{\Large{Abdelrahman Ramadan}}

\maketitle

\chapter*{Abstract}
Wildfires pose a severe challenge to ecosystems and human settlements, exacerbated by climate change and environmental factors. Traditional wildfire modeling, while useful, often fails to adapt to the rapid dynamics of such events. This report introduces the (Wildfire Autonomous Response and Prediction Using Cellular Automata) WARP-CA model, a novel approach that integrates terrain generation using Perlin noise with the dynamism of Cellular Automata (CA) to simulate wildfire spread. We explore the potential of Multi-Agent Reinforcement Learning (MARL) to manage wildfires by simulating autonomous agents, such as UAVs and UGVs, within a collaborative framework. Our methodology combines world simulation techniques and investigates emergent behaviors in MARL, focusing on efficient wildfire suppression and considering critical environmental factors like wind patterns and terrain features.

\tableofcontents

\listoffigures

\chapter{Introduction}
\section{Background}
Wildfires have become one of the most challenging natural disasters to predict, control, and mitigate. Driven by a combination of climatic, environmental, and anthropogenic factors, their unpredictable nature and the devastating impact they have on ecosystems and human settlements necessitate the development of advanced modeling techniques and control strategies \cite{tymstra_wildfire_2020, noauthor_canadian_2023, noauthor_international_nodate, news_wildfires_nodate}. 

Regions such as Canada have been particularly affected, witnessing a surge in wildfire activity. The urgency of the situation is further highlighted by the vast amounts of resources required for fire suppression, loss of habitats, and the significant economic repercussions for affected communities.

\section{Existing Techniques and Their Limitations}
Traditional wildfire simulations have utilized a range of methods, from the well-known Rothermel model calibrated with genetic algorithms \cite{pereira_review_2022}, to the more recent techniques that harness the power of complex network modeling \cite{perestrelo_modelling_2021}. Machine learning has also seen its application in this realm, with methods such as the Least-Squares Support-Vector Machines (LSSVM) \cite{xu2022modeling} combined with Cellular Automata (CA) \cite{byari_multi-scale_2022}. However, while these methods have made significant strides in wildfire modeling, they still fall short in dynamically responding to the fast-paced changes that occur during a wildfire event. 

\section{Need for a Novel Approach}
The integration of adaptive network growth models provides a promising avenue, allowing models to offer dynamic responses to varying conditions. Furthermore, the potential of incorporating autonomous agents such as Unmanned Aerial Vehicles (UAVs) and Unmanned Ground Vehicles (UGVs) within a Reinforcement Learning (RL) framework offers a tangible method for actionable control strategies during a wildfire event. However, as models scale up in complexity, they also introduce emergent behaviors, which, while offering sophisticated strategies, can introduce challenges due to their unpredictability \cite{liu2019emergent}.

Given this context, the objective of this report is to delve into the development and application of the WARP-CA model, exploring its potential in addressing the aforementioned challenges and contributing to the global efforts in wildfire management.

\section{Contributions}
This project introduces a novel approach in the realm of wildfire simulation and management, with a particular focus on the integration of world simulation techniques and the exploration of emergent behaviors through Multi-Agent Reinforcement Learning (MARL). The key contributions of this work are as follows:

\begin{itemize}
    \item \textbf{Integration of World Simulation Techniques:} Combined use of Perlin Noise for terrain generation and CA for fire spread simulation, providing a more comprehensive and nuanced approach to modeling wildfire dynamics.
    \item \textbf{Exploration of Emergent Behaviors in MARL:} Application of MARL to investigate how interactions among multiple autonomous agents, such as UAVs and UGVs, can lead to emergent behaviors that enhance wildfire suppression and forest conservation.
    \item \textbf{Focus on Efficient Wildfire Suppression:} Implementation of SARL and MARL frameworks to develop and test strategies that potentially accelerate wildfire suppression, aiming to minimize ecological damage and resource expenditure.
    \item \textbf{Incorporation of Environmental Factors:} Accounting for critical environmental variables such as wind patterns, vegetation types, and terrain features in the simulation to better understand their impact on fire behavior.
\end{itemize}

These contributions represent focused advancements in the field of wildfire simulation, specifically in the areas of integrated world simulation and the application of MARL to study and leverage emergent behaviors for more effective wildfire management.

\chapter{Literature Review}

\section{Comprehensive Review on Wildfire Simulation: Models, Algorithms, and Terrain Generation}

This section delves into the dynamic field of wildfire simulation, CA, Rothermel Model, Genetic Algorithms, and Perlin Noise-based terrain generation, and their integration in simulating forest fire dynamics.

\textbf{Wildfire simulation has matured into a multifaceted domain, amalgamating diverse models to replicate the intricate behavior of fire spread.} The Rothermel Model, pivotal in predicting fire behavior, underpins many modern simulation methods. It calculates the rate of fire spread based on fuel properties, topographical features, and meteorological conditions, thus providing a foundational understanding of fire dynamics \cite{yin_simulation_2018, zhang_study_2021, wahyuni_investigating_2021}. Integrating Genetic Algorithms (GAs) into wildfire simulations has marked a significant leap in enhancing predictive accuracy and adaptability. By emulating evolutionary processes, GAs refine parameters in simulation models, bolstering their adaptability to diverse environmental contexts. This technique has shown effectiveness in customizing models like the Rothermel Model for local conditions, thereby augmenting their forecasting accuracy \cite{pereira_review_2022}.

\textbf{Recent research has highlighted Perlin Noise as a novel terrain generation method in world simulations, contributing to the creation of realistic and varied landscapes.} This technique, utilizing a gradient noise function, crafts terrains that significantly influence fire behavior and spreading patterns, aligning with efforts to include intricate environmental elements in fire dynamics models \cite{fischer_autobiomes_2020, mastorakos_hybrid_2022}.

Our methodology, centered on Perlin Noise for world simulation, presents a pioneering stride in this field. Segmenting the generated Perlin noise into distinct environmental categories allows for a nuanced representation of diverse landscapes, crucial for comprehending fire spread across varied terrains \cite{le_procedural_nodate, jain_adaptive_2023}. Our approach, integrating sophisticated terrain modeling in fire simulations, significantly advances the replication of real-world fire scenarios \cite{rosadi_prediction_2021, zhao_simulation_2021, sun_adaptive_2021}.

\textbf{An enhanced approach to wildfire simulation entails incorporating comprehensive environmental data, such as forest maps, terrain elevation, and historical weather patterns.} These datasets provide critical insights into tree types, fuel content, age, elevation, and weather metrics, imperative for precise wind modeling via finite state models and fluid mechanics. Resources like the Canadian Wildland Fire Information System (CWFIS) Datamart, Global Forest Watch Open Data Portal, Geographic Information Systems from Novascotia.ca, and others, offer valuable geospatial data instrumental in creating a rich simulation environment. This approach aligns with sophisticated models and calculation systems for studying wildland fire behavior \cite{canada_forest_2013, noauthor_global_nodate, noauthor_geographic_nodate, noauthor_canadas_nodate, noauthor_topoview_nodate, noauthor_canada_nodate, pastor2003mathematical, forthofer2007modeling}.

\textbf{In conclusion, wildfire simulation is a rapidly evolving domain, with the advent of advanced models like the Adaptive CA, Rothermel Model, and Genetic Algorithms playing crucial roles.} Our research methodology, especially in terrain generation using Perlin Noise and wildfire spread simulation using CA, signifies a substantial contribution to this field. It aligns with and extends current research by introducing new layers of complexity and realism in simulating wildfire dynamics \cite{byari_multi-scale_2022, jain_adaptive_2023, bhakti_fire_2020}. Our literature review uncovered a lack of studies employing these specific world simulation techniques for wildfire simulation.

\section{Autonomous Wildfire Management and Response Systems: A MARL Perspective}

MARL extrapolates the principles of single-agent reinforcement learning to scenarios with multiple interacting agents. In MARL, each agent operates within Markov Decision Processes (MDPs), with their actions intricately influencing the dynamics of states and rewards. This concept is particularly pertinent in wildfire management, where a myriad of entities such as firefighters, drones, and environmental factors interact in a shared ecosystem \cite{shapley1953stochastic, littman1994markov}. Markov Games, in this context, model the interplay between multiple agents, whose collective decisions and actions determine state transitions and rewards. In wildfire management, these games can simulate various firefighting teams or autonomous systems operating in tandem or contention \cite{littman1994markov, basar1999dynamic}.

\textbf{Cooperative MARL settings}, where agents unite towards a common objective like wildfire suppression, can be conceptualized as multi-agent MDPs (MMDPs) or team Markov games. Here, the collective effort is channeled towards a unified reward function \cite{boutilier1996planning, wai2018multi}. In competitive MARL settings, such as zero-sum games, agents' goals are diametrically opposed, akin to competing land management agencies in wildfire scenarios. This setting is reflective of scenarios where each party's gain is another's loss \cite{littman1994markov, silver2017mastering}.

\textbf{The mixed setting in MARL, or general-sum games, aptly represents wildfire management scenarios with overlapping yet distinct objectives among diverse groups.} This setting accommodates the complexities and nuances of real-world situations \cite{hu2003nash, OpenAI_dota}. Extensive-form games adeptly handle scenarios with imperfect information, a common hurdle in wildfire management. Agents' decision-making is based on limited or uncertain environmental information, mirroring the unpredictability inherent in real-world wildfires \cite{osborne1994course, heinrich2015fictitious}.

Several studies exemplify MARL applications in wildfire management:

\begin{itemize}
    \item \textbf{Collaborative Auto-Curricula}: Siedler (2022) investigates a MARL system with a Graph Neural Network communication layer for efficient resource distribution in wildfire management, underscoring the value of agent collaboration \cite{siedler_collaborative_2022}.
    \item \textbf{UAVs and Deep Q-Learning}: Viseras et al. (2021) explore using UAVs for wildfire front monitoring, employing deep Q-learning to underscore the potential of autonomous aerial vehicles in fire tracking \cite{viseras_wildfire_2021}.
    \item \textbf{Distributed Deep Reinforcement Learning}: Haksar and Schwager (2018) present distributed deep reinforcement learning for maneuvering aerial robots to combat forest fires, demonstrating decentralized control's effectiveness in intricate environments \cite{haksar_distributed_2018}.
    \item \textbf{Swarm Navigation and Control}: Ali et al. (2023) concentrate on distributed multi-agent deep reinforcement learning for UAV swarm navigation in wildfire monitoring, showcasing advanced coordination techniques \cite{ali_distributed_2023}.
\end{itemize}

Integrating MARL into wildfire management is pivotal in developing robust, autonomous systems for effective wildfire response. MARL's complex interaction models lay the theoretical groundwork for designing advanced simulation algorithms and optimizing strategies for wildfire prevention, control, and mitigation \cite{shapley1953stochastic, littman1994markov}. Future research in this arena could focus on devising sophisticated algorithms that bolster collaboration among autonomous agents, enhance decision-making amidst uncertainty, and support real-time adaptive strategies. Utilizing MARL can pave the way for more efficacious systems in wildfire prediction, monitoring, and response \cite{osborne1994course, heinrich2015fictitious}.

\chapter{Methodology}
\section{World Simulation}

\subsection{Terrain Generation with Perlin Noise}
Perlin noise, a gradient noise function, is central to our approach for generating heterogeneous terrain in forest fire dynamics simulation. The mathematical basis of Perlin noise, its parameters, and application in terrain generation are detailed below.

\subsubsection{Mathematical Definition of Perlin Noise}
Introduced by Ken Perlin, Perlin noise is a gradient noise function that generates coherent noise ideal for simulating natural patterns. The process involves several steps and parameters, detailed as follows:

\paragraph{Gradient Vector Generation}
At each point \( P_i \) on a lattice, a pseudo-random gradient vector \( \vec{G}_i \) is generated. The distribution and orientation of these vectors significantly influence the characteristics of the resulting noise pattern.

\paragraph{Noise Calculation}
For a point \( \vec{x} \) in space, the Perlin noise value, \( N(\vec{x}) \), is calculated by:
\begin{enumerate}
    \item Identifying the unit cube that encloses \( \vec{x} \).
    \item Computing the dot product \( \vec{G}_i \cdot \vec{D}_i \) for each of the cube's corners, where \( \vec{D}_i = \vec{x} - \vec{P}_i \) is the vector from the corner \( \vec{P}_i \) to \( \vec{x} \).
    \item Applying a smoothing function to interpolate these dot products, yielding the final noise value.
\end{enumerate}

\paragraph{Smoothing Function}
The smoothing function is typically a polynomial, such as the quintic function \( f(t) = 6t^5 - 15t^4 + 10t^3 \), which ensures the continuity and smoothness of the noise pattern. Which is illustrated below in Figure \ref{fig:smooth_perlin}

\begin{figure}[H]
    \centering
    \includegraphics{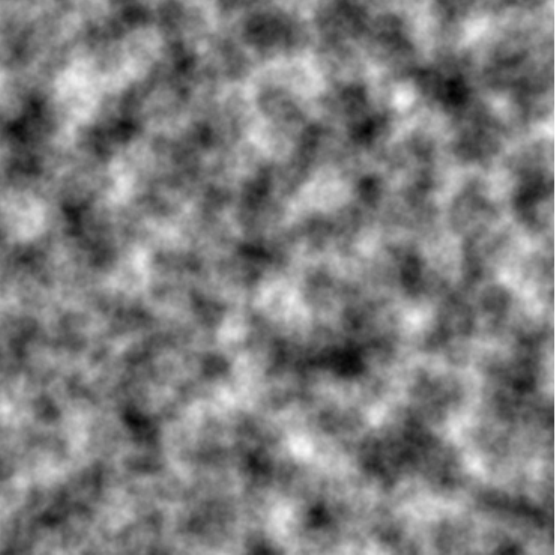}
    \caption{Smoothed Perlin Noise Map}
    \label{fig:smooth_perlin}
\end{figure}
\paragraph{Parameters of Perlin Noise}
Several parameters control the characteristics of the Perlin noise:
\begin{itemize}
    \item \textbf{Scale (\( \sigma \))}: Controls the granularity of the noise. Smaller values lead to finer details, while larger values create broader patterns.
    \item \textbf{Octaves (\( \omega \))}: Determines the number of layers of noise. Higher octaves add complexity and finer details.
    \item \textbf{Persistence (\( \rho \))}: Influences the amplitude of each octave. Higher persistence values emphasize finer details.
    \item \textbf{Lacunarity (\( \lambda \))}: Adjusts the frequency of each octave. Higher lacunarity increases the frequency, adding more small-scale details.
\end{itemize}

The final function form representing the Perlin noise surface is a composite of multiple layers of noise, each with its own scale, persistence, and lacunarity. Mathematically, it can be described as a sum of successive noise functions, where each function \( \text{noise}_i(x, y) \) is scaled and its amplitude adjusted according to the given parameters. The form can be represented as:

\[ f(x, y) = \sum_{i=0}^{\omega} \rho ^i \cdot \text{noise}_i\left(\frac{x}{\sigma  \cdot \lambda^i}, \frac{y}{\sigma \cdot \lambda^i}\right) \]

This formula combines the effects of scale, octaves, persistence, and lacunarity to generate a complex, layered noise pattern.

\subsubsection{Application in Terrain Generation}
Terrain generation using Perlin noise involves mapping the noise values to different terrain types. This process is crucial for simulating realistic and varied terrains, impacting the spread and behavior of wildfires. The mapping can be described as follows:

\begin{itemize}
    \item Assign noise value ranges to specific terrain types (e.g., mountains, plains, forests).
    \item Use these classifications to create a diverse landscape that realistically affects fire dynamics.
\end{itemize}

\subsubsection{Advantages and Suitability for Fire Dynamics Simulation}
Perlin noise is particularly suitable for forest fire dynamics simulations due to:
\begin{itemize}
    \item \textbf{Natural Appearance:} It creates terrains that closely mimic real-world landscapes.
    \item \textbf{Control and Predictability:} Adjustable parameters allow for controlled variation in terrain features.
    \item \textbf{Efficiency:} Generates complex landscapes with relatively low computational overhead.
\end{itemize}

These characteristics make Perlin noise an ideal choice for generating terrains in simulations aimed at studying the complex behavior of wildfires, as they provide a realistic and varied environment that significantly influences fire spread patterns.

\paragraph{1. Environmental Categorization}

The generated Perlin noise is segmented into four environmental categories, each corresponding to a unique terrain type:
\begin{itemize}
    \item Lake: Noise values less than the lake threshold.
    \item Wetland: Noise values between lake and wetland thresholds.
    \item Grassland: Noise values between wetland and grassland thresholds.
    \item Forest: Noise values greater than or equal to the grassland threshold.
\end{itemize}

\begin{figure}[htbp]
    \centering
    \begin{subfigure}{.45\textwidth}
        \centering
        \includegraphics[width=\linewidth]{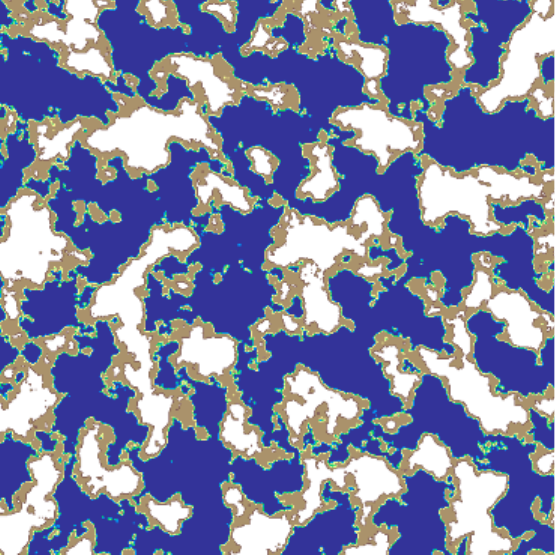}
        \caption{Scale 50}
    \end{subfigure}
    \hfill
    \begin{subfigure}{.45\textwidth}
        \centering
        \includegraphics[width=\linewidth]{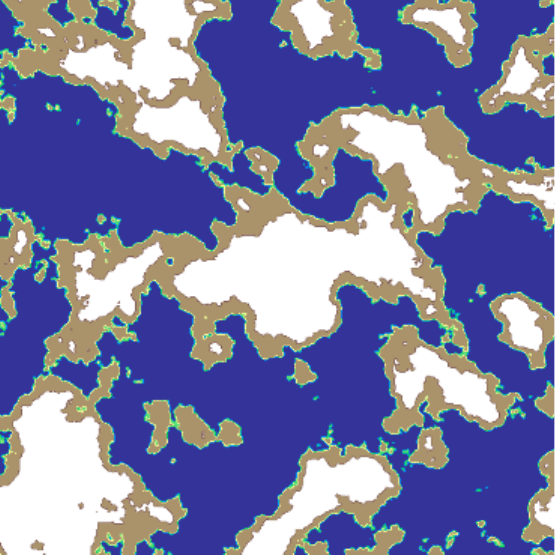}
        \caption{Scale 100}
    \end{subfigure}
    \par\bigskip
    \begin{subfigure}{.45\textwidth}
        \centering
        \includegraphics[width=\linewidth]{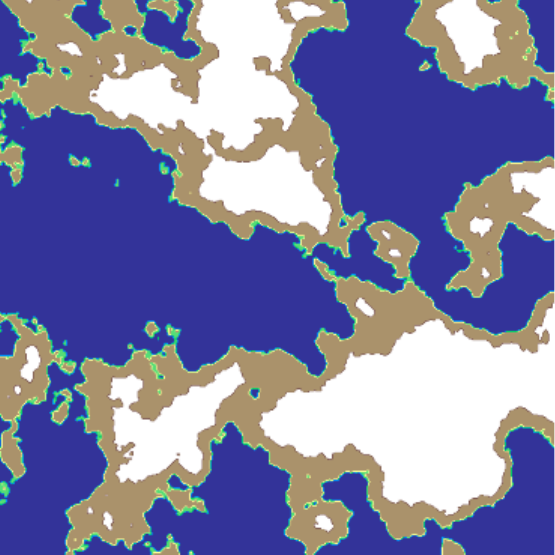}
        \caption{Scale 150}
    \end{subfigure}
    \hfill
    \begin{subfigure}{.45\textwidth}
        \centering
        \includegraphics[width=\linewidth]{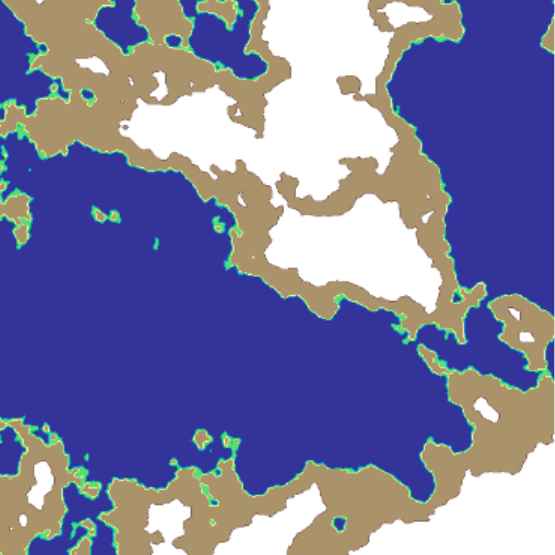}
        \caption{Scale 200}
    \end{subfigure}
    \caption{The Effect of Choosing Different Scales on Perlin Noise Generation.}
    \label{fig:perlin_noise_scales}
\end{figure}

In Figure \ref{fig:perlin_noise_scales} we can observe how the tuning of one parameter $\sigma$ can affect the generated map drastically, changing $\sigma$ can amount to zooming in and out on a map as can be noticed in Figure \ref{fig:perlin_noise_scales}.

\subsection{Wildfires Fire Spread Simulation using CA}
\subsubsection{Model Description}

This sub-section presents a detailed cellular automaton model for simulating forest fires, encompassing the dynamics of forest growth, fire spread, burning, and regeneration.

\paragraph{Parameters}
\begin{itemize}
    \item $N_x, N_y$: Dimensions of the forest grid.
    \item $S_{\text{max}}$: Maximum growth state of a tree (typically 10).
    \item $F_{\text{max}}$: Number of fire and ash states (typically 5, states 11 to 15).
    \item $\alpha$: Probability of fire spreading to adjacent cells.
    \item $\beta$: Probability of spontaneous tree ignition.
    \item $\tau$: Growth rate of trees.
    \item $A_{\text{min}}$: Minimum age for trees to burn.
\end{itemize}

\subsubsection{Cellular Automaton Rules}
\begin{enumerate}
    \item \textbf{Initial Forest State}: The initial state of the forest is generated randomly based on the maximum growth state of trees.
    \begin{equation}
        \text{forest} = \text{round}(S_{\text{max}} \times \text{rand}(N_x, N_y))
    \end{equation}

    \item \textbf{Tree Growth}: Trees grow according to a growth probability, which is a function of their current state and a growth rate parameter $\tau$.
    \begin{equation}
        p_{\text{Growth}} = \exp\left(-\frac{\text{forest}(\text{treeIdx})}{S_{\text{max}}} \times \tau\right)
    \end{equation}

    \item \textbf{Fire Dynamics}: Fire dynamics are modeled as a multi-step process, including ignition, burning stages, and transition to ash.
    \begin{enumerate}
        \item \textbf{Ignition}: Trees can spontaneously ignite based on a probability $\beta$, which is influenced by their growth state.
        \begin{equation}
            p_{\text{Ignite}} = \beta \times \frac{\exp\left(\frac{\text{forest}(\text{treeIdx})}{S_{\text{max}}} \times 10\right)}{\exp(10)} \times \frac{1}{N_x \times N_y}
        \end{equation}
        \item \textbf{Burning Stages}: Trees transition through burning stages (states 11 to 15) before turning into ash.
        \item \textbf{Transition to Ash}: After burning, trees transition to an ash state (state 0).
    \end{enumerate}

    \item \textbf{Regeneration}: Ash cells have the potential to grow new trees, representing the regeneration of the forest.
\end{enumerate}

\begin{figure}[htbp]
    \centering
    \begin{subfigure}{.45\textwidth}
        \centering
        \includegraphics[width=\linewidth]{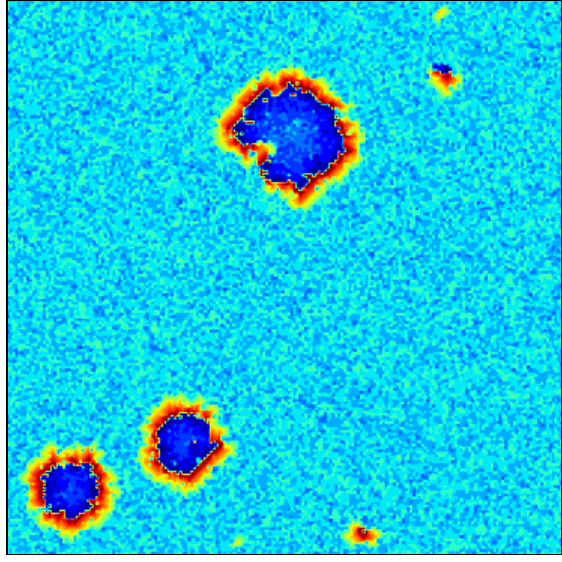}
        \caption{Frame 440}
    \end{subfigure}
    \hfill
    \begin{subfigure}{.45\textwidth}
        \centering
        \includegraphics[width=\linewidth]{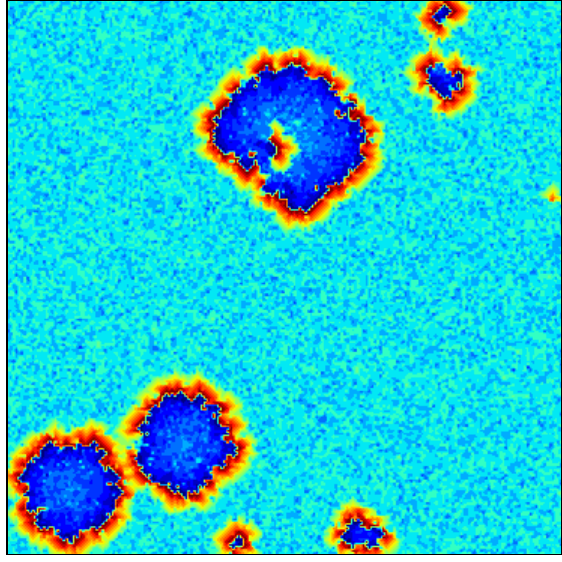}
        \caption{Frame 450}
    \end{subfigure}
    \par\bigskip
    \begin{subfigure}{.45\textwidth}
        \centering
        \includegraphics[width=\linewidth]{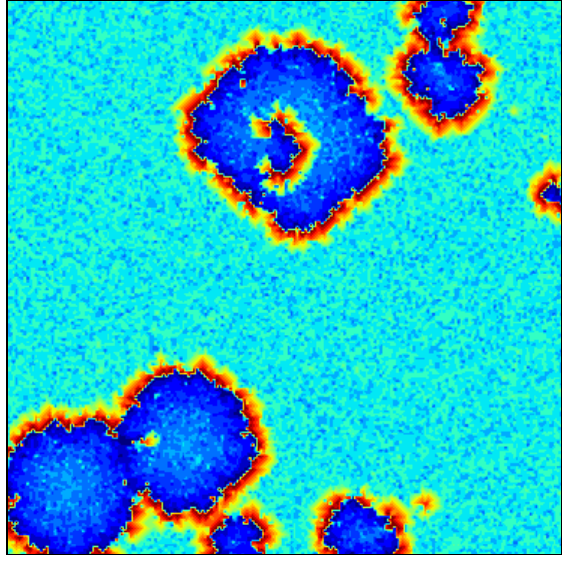}
        \caption{Frame 460}
    \end{subfigure}
    \hfill
    \begin{subfigure}{.45\textwidth}
        \centering
        \includegraphics[width=\linewidth]{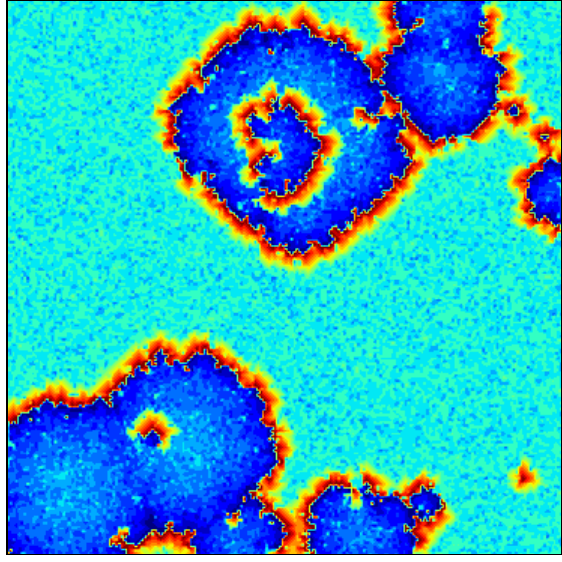}
        \caption{Frame 470}
    \end{subfigure}
    \caption{Progression of Fire Spread using CA}
    \label{fig:fire_progress}
\end{figure}

We can observe how the fire spreads across multiple frames of simulation as shown in Figure \ref{fig:fire_progress}, where an emergent behavior exhibiting what would amount to a simulation of fire spreading in a gird world just by defining our grid and dictating some simple rules that govern the interactions between these grid cells.

\subsubsection{Fire Propagation and Environmental Factors}
Fire propagation is influenced by multiple environmental factors, including terrain, wind, and vegetation type. These factors affect the likelihood of fire spread from one cell to its neighbors. Different vegetation types (forest, grassland, wetland) have distinct probabilities of catching and propagating fire. These probabilities are adjusted based on environmental conditions like slope and wind.

\paragraph{Slope Calculation}
The slope between two cells affects fire propagation. It is calculated based on the height difference between neighboring cells.
\begin{equation}
    \text{slope} = \arctan\left(\frac{\text{height difference}}{\text{distance between cells}}\right)
\end{equation}

\subsubsection{Wind Influence in Fire Spread}
The wind model in our simulation represents a simplified version of the more complex wind behavior in wildland fire spread, as discussed in Forthofer's research. This simplification is necessary for computational efficiency and ease of implementation in our cellular automaton framework.

\paragraph{Wind Model Used in Simulation}
Our model assigns a basic wind vector across the simulation grid. The wind vector is defined as:
\begin{equation}
    \vec{W} = \begin{bmatrix}
           U \\
           V
         \end{bmatrix}
\end{equation}
where \( U \) and \( V \) are the wind velocity components in the x and y directions, respectively. 

\paragraph{Custom Wind Model}
The wind velocity components used in the simulation are defined by the following equations:
\begin{align}
    U &= -1 - X^2 + X + Y \\
    V &= \frac{1 + X^2 - Y^2}{10}
\end{align}
where \( X \) and \( Y \) represent the spatial coordinates on the simulation grid.
The wind direction is computed as:
\begin{equation}
    \theta_{\text{wind}} = \arctan2(V, U)
\end{equation}

\paragraph{Comparison with Advanced Models}
In contrast, Forthofer's study utilizes complex Computational Fluid Dynamics (CFD) and mass-consistent models for accurate wind simulations in varied terrains. These models account for microscale wind variations and are more precise but computationally intensive.

\begin{figure}[htbp]
    \centering
    \begin{subfigure}{.45\textwidth}
        \centering
        \includegraphics[width=\linewidth]{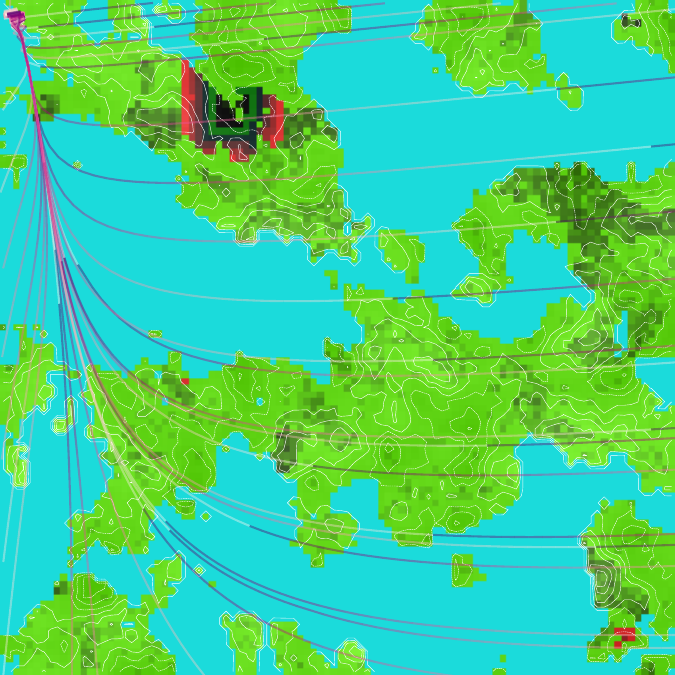}
        \caption{Frame 350}
    \end{subfigure}
    \hfill
    \begin{subfigure}{.45\textwidth}
        \centering
        \includegraphics[width=\linewidth]{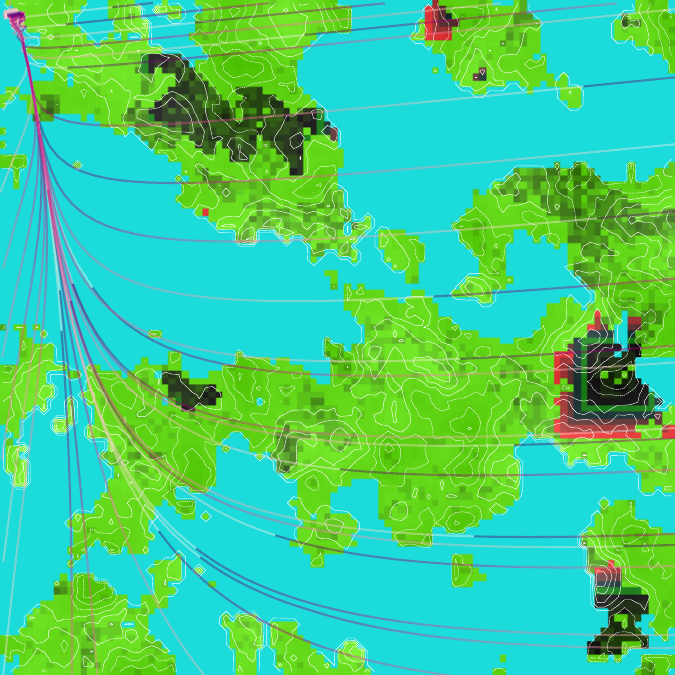}
        \caption{Frame 360}
    \end{subfigure}
    \par\bigskip
    \begin{subfigure}{.45\textwidth}
        \centering
        \includegraphics[width=\linewidth]{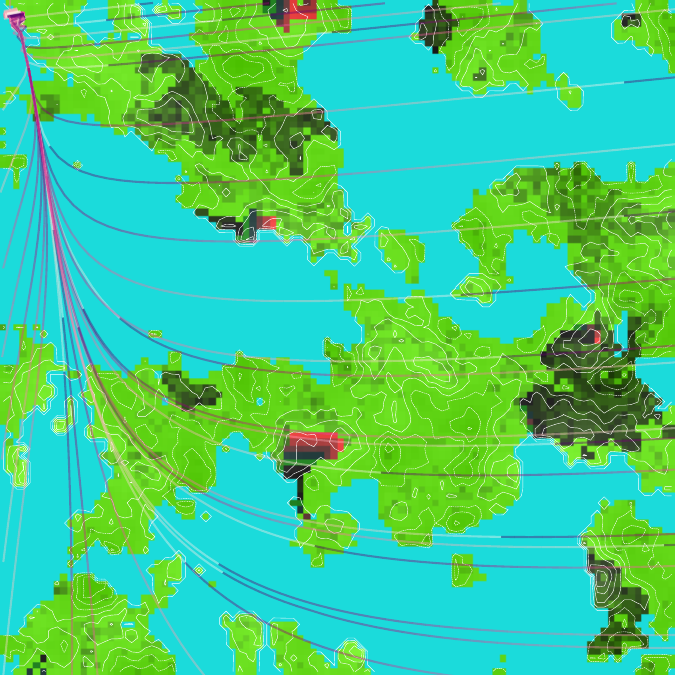}
        \caption{Frame 370}
    \end{subfigure}
    \hfill
    \begin{subfigure}{.45\textwidth}
        \centering
        \includegraphics[width=\linewidth]{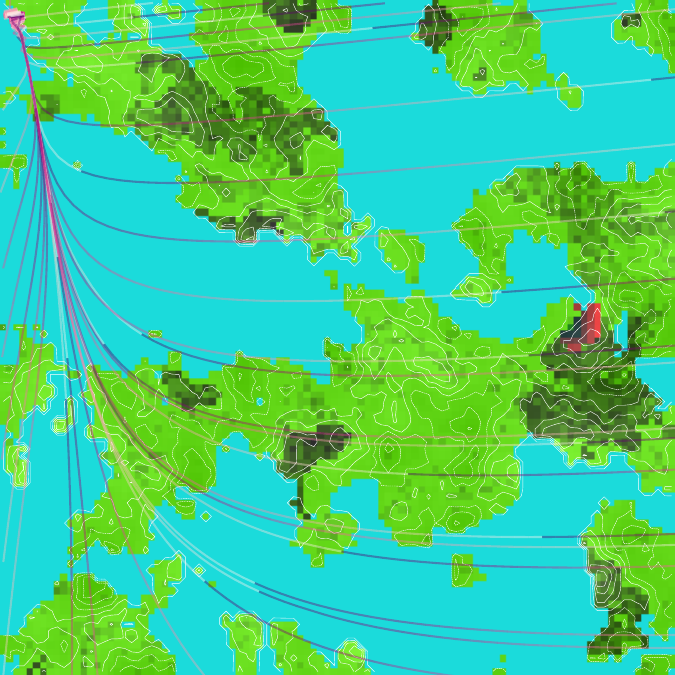}
        \caption{Frame 380}
    \end{subfigure}
    \caption{Progression of the Forest Fire simulation with additional Environmental Complexities}
    \label{fig:fire_simulation_stages}
\end{figure}

\paragraph{Rationale for Simplification}
Our simplified model provides a practical balance, incorporating the influence of wind on fire spread while maintaining computational efficiency. It allows us to capture the directional influence of wind on fire propagation without the computational burden of more complex models. This approach is effective for simulating general wind effects in fire behavior, particularly suitable for large-scale simulations where detailed wind patterns are less critical. Finally in Figure \ref{fig:fire_simulation_stages} we can see that we successfully integrated CA with Perlin noise-generated terrain, in a simulation environment influenced by environmental factors described in this section.

\begin{algorithm}[H]
\caption{Forest Fire Dynamics Simulation}
\begin{algorithmic}[1]
\Procedure{ForestFireSimulation}{}
    \State Initialize forest grid of size $N_x \times N_y$ with random tree ages
    \For{each simulation frame}
        \For{each cell in forest}
            \If{cell contains a tree}
                \State Compute $p_{\text{Growth}}$
                \If{random value < $p_{\text{Growth}}$}
                    \State Increase tree age
                \EndIf
            \EndIf
            \If{cell is on fire}
                \State Propagate fire with probability $\alpha$
                \State Update burning states
            \EndIf
            \If{cell contains a tree and is not on fire}
                \State Compute $p_{\text{Ignite}}$
                \If{random value < $p_{\text{Ignite}}$}
                    \State Ignite the tree
                \EndIf
            \EndIf
            \If{cell is in the final burning state}
                \State Transition to ash (state 0)
            \EndIf
            \If{cell is ash}
                \State Chance to regenerate a tree
            \EndIf
        \EndFor
        \State Save and display current state of forest
    \EndFor
\EndProcedure
\end{algorithmic}
\end{algorithm}

\section{Single Agent Reinforcement Learning (SARL) Framework}

\subsection{Environment Dynamics}
The RL Environment is defined as follows:

\begin{itemize}
    \item \textbf{State Space} ($S$): The state $s_t \in S$ is a flattened array representing the agent's local view, which includes terrain, tree, and fire states.
    \item \textbf{Action Space} ($A$): The action space is discrete with actions denoted as $A = \{0, 1, \dots, 9\}$, corresponding to specific movement and interaction actions.
    \item \textbf{Transition Dynamics}: The dynamics involve fire spread influenced by neighboring cells, wind, and terrain, as well as the agent's movement.
    \item \textbf{Reward Function}: Defined as $R(s_t, a_t, s_{t+1})$, it combines immediate action rewards with strategic considerations.
\end{itemize}

\subsection{Reinforcement Learning Algorithms}
We employ three distinct RL algorithms:

\begin{itemize}
    \item \textbf{Proximal Policy Optimization (PPO)}: Utilizes a policy $\pi_\theta(a|s)$, with the objective function:
    \[ L^{PPO}(\theta) = \mathbb{E}_t \left[ \min(r_t(\theta)\hat{A}_t, \text{clip}(r_t(\theta), 1-\epsilon, 1+\epsilon)\hat{A}_t) \right] \]
    where $r_t(\theta)$ is the probability ratio and $\hat{A}_t$ is the advantage estimate.
    
    \item \textbf{Advantage Actor-Critic (A2C)}: Employs separate actor and critic networks, with the loss function:
    \[ L^{A2C}(\theta) = -\log \pi_\theta(a_t|s_t) \hat{A}_t + \lambda (V_\theta(s_t) - R_t)^2 \]
    
    \item \textbf{Deep Q-Network (DQN)}: Uses a Q-network for state-action value estimation, with the loss function:
    \[ L^{DQN}(\theta) = \mathbb{E}_{(s,a,r,s') \sim \text{replay}} \left[ (r + \gamma \max_{a'} Q_{\theta^-}(s', a') - Q_\theta(s, a))^2 \right] \]
\end{itemize}

\subsection{Reward Function}
The reward function is structured to encourage strategic actions:

\begin{itemize}
    \item \textbf{Base Rewards}: Movement ($R_{\text{move}} = -0.2$), extinguishing fire, and clearing vegetation.
    \item \textbf{Strategic Actions}: Significant reward for creating a fire barrier ($R_{\text{barrier}} = 50$) and potential-based rewards.
    \item \textbf{Penalties}: For failed extinguishing attempts and unnecessary clearing.
    \item \textbf{Normalization}: Rewards are normalized to mitigate scaling issues.
\end{itemize}

The function $R(s, a, s') = R_{\text{base}}(a, s) + R_{\text{strategic}}(s, s') + R_{\text{potential}}(s, s')$ encapsulates these elements.

\subsection{Integration with Fire Dynamics Simulation}
In this environment, the agent's actions directly impact fire spread and forest preservation. The learning algorithms are tasked with optimizing strategies to minimize fire damage and maximize the effectiveness of the agent's actions in controlling and managing forest fires.

\section{Multi-Agent Reinforcement Learning (MARL) Framework}

\subsection{Defining the Environment}

\subsubsection{State Space}
In our wildfire management scenario, the state space for each agent $i$ (a fire extinguishing unit) at time $t$ is denoted as $s^i_t$. It includes factors like the agent's location, the status of the fire, and surrounding environmental conditions. The global state $S_t$ is then:
\[
S_t = \{s^1_t, s^2_t, \ldots, s^M_t\}
\]
where $M$ is the number of agents (firefighting units).

\subsubsection{Action Space}
The action space $A^i$ for each agent might include moving in different directions, extinguishing fire, or coordinating with other agents. The collective action space $A$ is:
\[
A = A^1 \times A^2 \times \ldots \times A^M
\]

\subsubsection{Observation Space}
Each agent observes $o^i_t$, reflecting its immediate surroundings and the state of the fire. The observation function is:
\[
O^i: S \times A \to O^i
\]

\subsubsection{Reward Function}
The reward function is critical in guiding agents towards effective fire management. It is defined as:
\[
R(S_t, A_t, S_{t+1}) = \sum_{i=1}^{M} R^i(s^i_t, a^i_t, s^i_{t+1})
\]
$R^i$ rewards actions like extinguishing fire or saving trees and penalizes harmful actions.

\subsection{Applying PPO in MARL for Wildfire Management}

\subsubsection*{Centralized Training with Decentralized Execution (CTDE)}

\paragraph*{Training Phase}
During training, a centralized approach is used where the joint policy $\Pi = \{\pi^1, \pi^2, \ldots, \pi^M\}$ is optimized using global information.

\paragraph*{Execution Phase}
Each agent acts independently based on its policy $\pi^i$ and local observation $o^i_t$.

\subsubsection{PPO Loss Function}
The PPO loss function in the context of MARL for our wildfire scenario is adapted to optimize the joint policy $\Pi$. The loss function for PPO in this setting, considering a simplified view, is:
\[
L^{PPO}(\theta) = \mathbb{E}_t \left[ \min(r_t(\theta)\hat{A}_t, \text{clip}(r_t(\theta), 1-\epsilon, 1+\epsilon)\hat{A}_t) \right]
\]
where $r_t(\theta)$ is the probability ratio of the new policy to the old policy, $\hat{A}_t$ is the advantage estimate at time $t$, and $\epsilon$ is a hyperparameter.

In MARL for firefighting, the advantage $\hat{A}_t$ for each agent is calculated considering both individual and collective performance in managing the fire and preserving the environment. The policy is updated to maximize this loss function, guiding agents to learn cooperative strategies for effective wildfire management.

\begin{figure}[htbp]
\centering
\includegraphics[width=\textwidth]{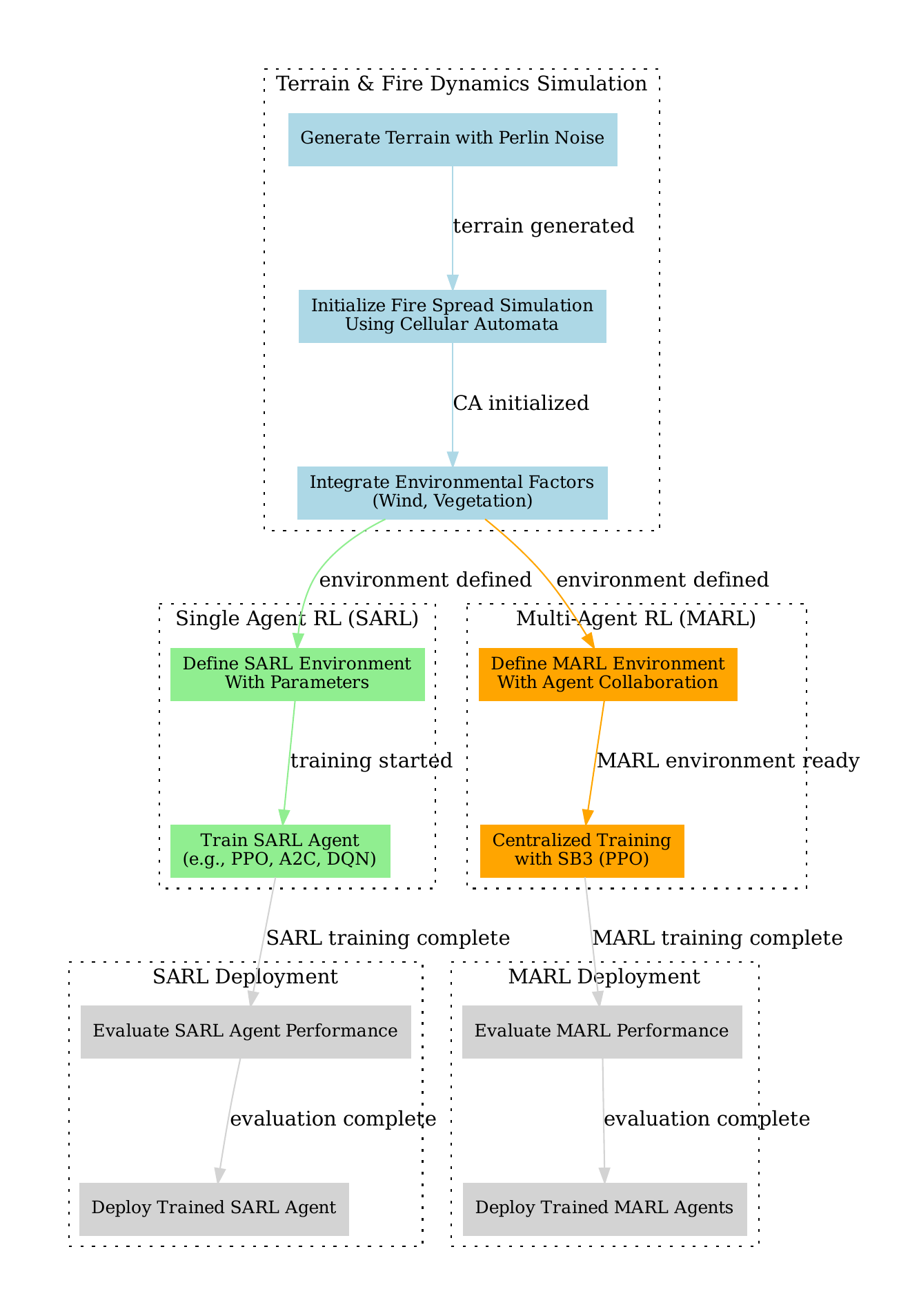}
\caption{Pipeline from World Simulation to SARL and MARL Training and Deployment.}
\label{fig:pipeline}
\end{figure}

The overall pipeline from terrain generation through world simulation to both SARL and MARL training and deployment is summarized in Figure \ref{fig:pipeline}. The process begins with the generation of terrain using Perlin Noise, followed by the initialization of fire spread simulation using Cellular Automata, and the integration of environmental factors. Two parallel streams then proceed independently: one for SARL and one for MARL. Each stream involves defining the environment, training the agents using suitable algorithms (such as PPO, A2C, DQN for SARL, and SB3 with PPO for MARL), and finally deploying the trained agents after performance evaluation.

\chapter{Simulation Results and Discussion}

\section{Simulation Environment}
In our setup, we have developed a custom simulation for a fire extinguishing scenario, represented as a grid where an agent can move and perform actions such as extinguishing fire or clearing vegetation. The environment is stochastic, with elements such as wind and terrain influencing the behavior of fires. The following subsections provide detailed descriptions of the environment and the hyperparameters used in the simulations.

\section{Simulation Machine Specifications}
Simulations were conducted on a dedicated machine to ensure consistent and reliable computational performance. The specifications of the machine used for the simulations are detailed below:

\begin{verbatim}
\footnotesize
Operating System: Ubuntu 20.04.6 LTS
CPU: 11th Gen Intel® Core™ i7-11800H @ 2.30GHz × 16 
GPU: NVIDIA GeForce RTX 3050
RAM: 32 GiB
Disk: 1 TB
\end{verbatim}

\section{Hyperparameters Definition}
Based on the methodology outlined, the hyperparameters for the terrain generation using Perlin noise and the cellular automaton model for wildfire spread simulation are defined as follows:

\begin{table}[ht]
\centering
\footnotesize 
\renewcommand{\arraystretch}{0.8} 
\begin{tabular}{l|c|l}
\toprule
\textbf{Hyperparameter} & \textbf{Symbol} & \textbf{Value} \\
\midrule
Scale & \( \sigma \) & 15 \\
Octaves & \( \omega \) & 6 \\
Persistence & \( \rho \) & 0.5 \\
Lacunarity & \( \lambda \) & 2 \\
Lake Threshold & $\text{Lake}_{\text{Th}}$ & 0.01 \\
Wetland Threshold & $\text{Wetland}_{\text{Th}}$ & 0.2 \\
Grassland Threshold & $\text{GrassL}_{\text{Th}}$ & 0.01 \\
Maximum Tree Growth State & \( S_{\text{max}} \) & 10 \\
Number of Fire States & \( F_{\text{max}} \) & 4 \\
Probability of Fire Spread & \( \alpha \) & 0.8, 1, 0.1 (forest, grassland, wetland) \\
Probability of Tree Ignition & \( \beta \) & 0.65 \\
Tree Growth Rate & \( \tau \) & 5 \\
Minimum Burning Age & \( A_{\text{min}} \) & 2 \\
\bottomrule
\end{tabular}
\caption{Hyperparameters for terrain generation and wildfire spread simulation.}
\label{table:hyperparameters_simulation}
\end{table}

\section{SARL Results}

\subsection{Environment Description}
The environment, termed \texttt{FireExtinguishingEnv}, is a grid world where the agent's objective is to control and extinguish fires. The state of each cell in the grid represents different types of terrain and fire intensity. The dynamics of fire spread are influenced by various factors including the type of vegetation, wind, and the presence of water bodies.

\subsection{Agent and Model Training}
The agent is trained using various reinforcement learning algorithms including PPO, A2C, DQN, DDPG, and SAC. The training process involves the agent interacting with the environment, receiving observations, and taking actions that are governed by a policy network. The policy network is updated using gradient ascent on expected returns. The Training process and agent interactions with the environment is shown in Figure \ref{flow:fire_ext_env_workflow_single}.

The following code snippet shows the instantiation and training of the agent using the PPO algorithm:

\begin{verbatim}
model = PPO('MlpPolicy', env, verbose=1, tensorboard_log=unique_log_dir,
device=device)
model.learn(total_timesteps=timesteps)
\end{verbatim}

\begin{figure}[h]
    \centering
    \includegraphics[width=\textwidth]{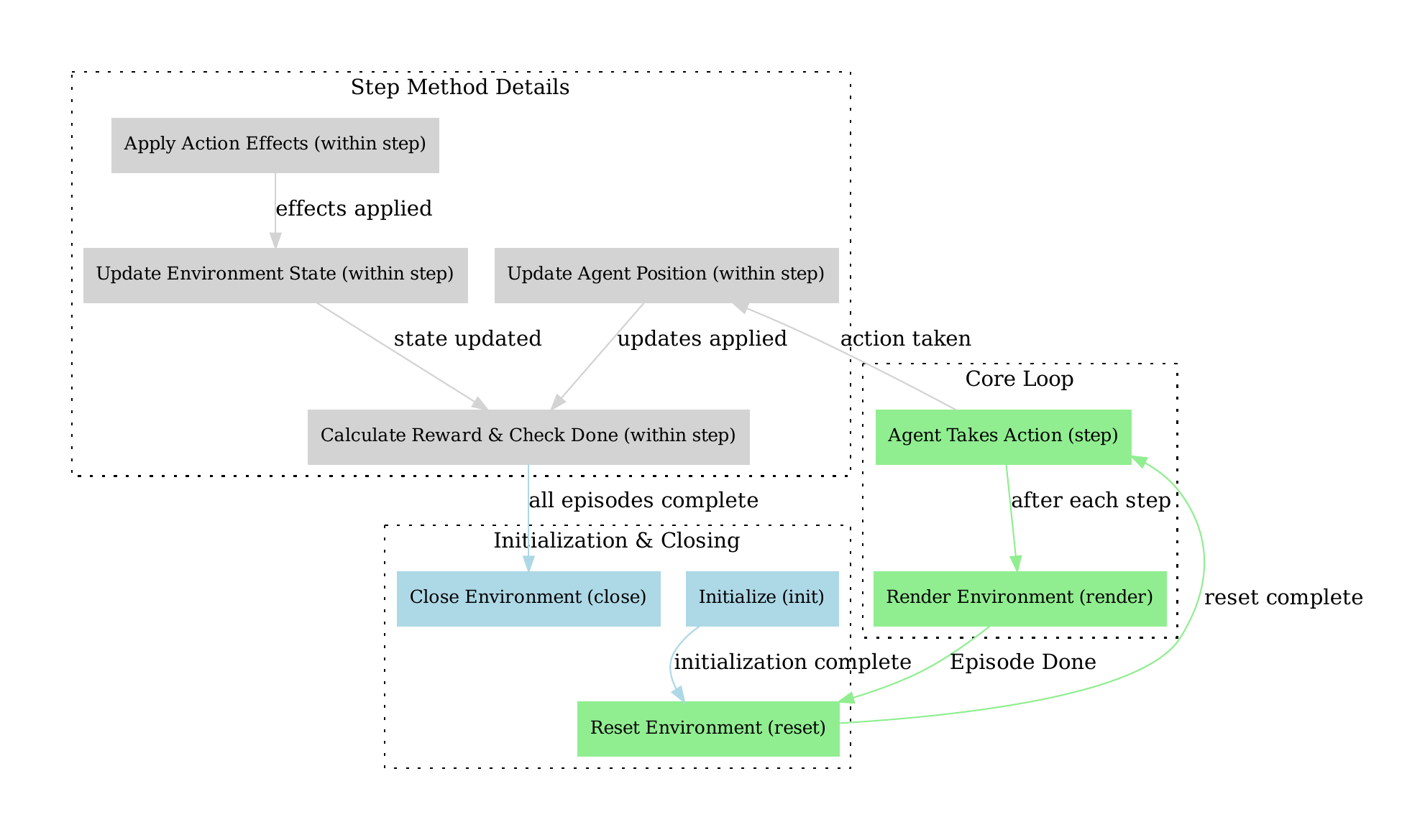}
    \caption{Training Flow of the Fire Extinguishing Environment with for a Single Agent.}
    \label{flow:fire_ext_env_workflow_single}
\end{figure}

\subsection{Performance Metrics}

This section provides an analysis of the training performance of our SARL environment using various reinforcement learning models, specifically PPO, A2C, and DQN. Performance metrics such as frame rate per second, loss, KL divergence, and others are considered.

\begin{figure}[H]
\centering
\includegraphics[width=0.7\textwidth]{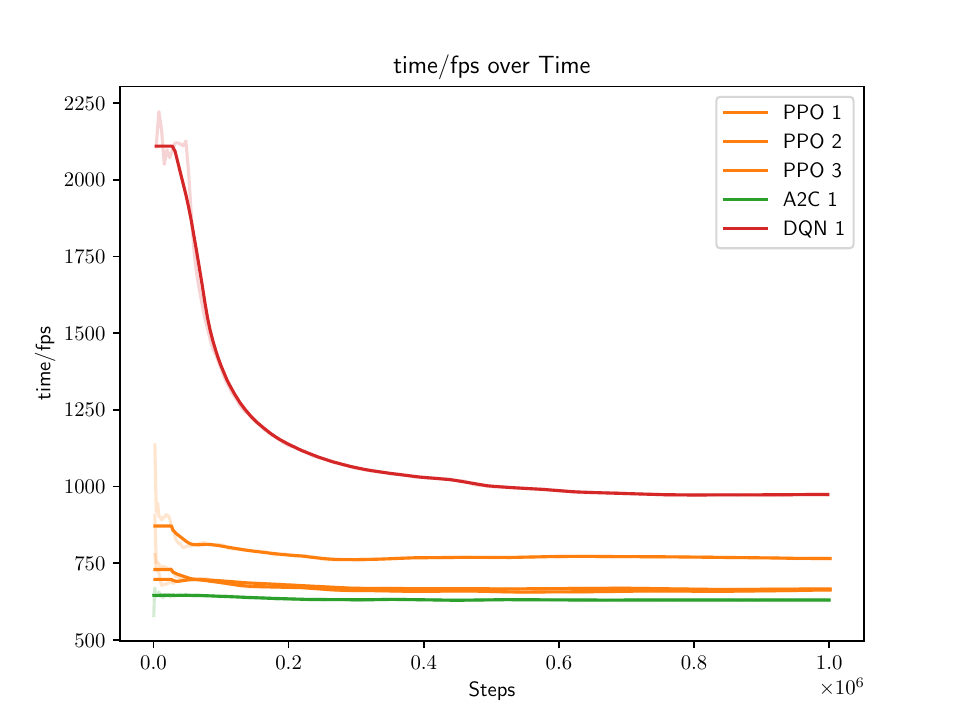}
\caption{Frame rate per second over time for different models.}
\label{fig:time_fps}
\end{figure}

\begin{figure}[H]
\centering
\includegraphics[width=0.7\textwidth]{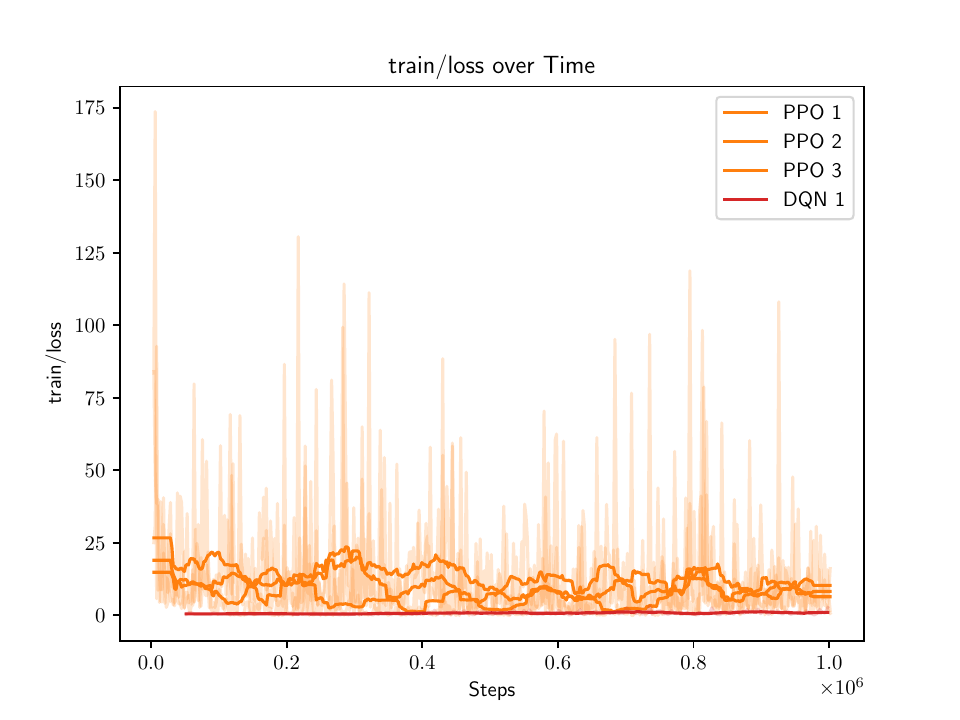}
\caption{Training loss over time for different models.}
\label{fig:train_loss}
\end{figure}

\begin{figure}[H]
\centering
\includegraphics[width=0.7\textwidth]{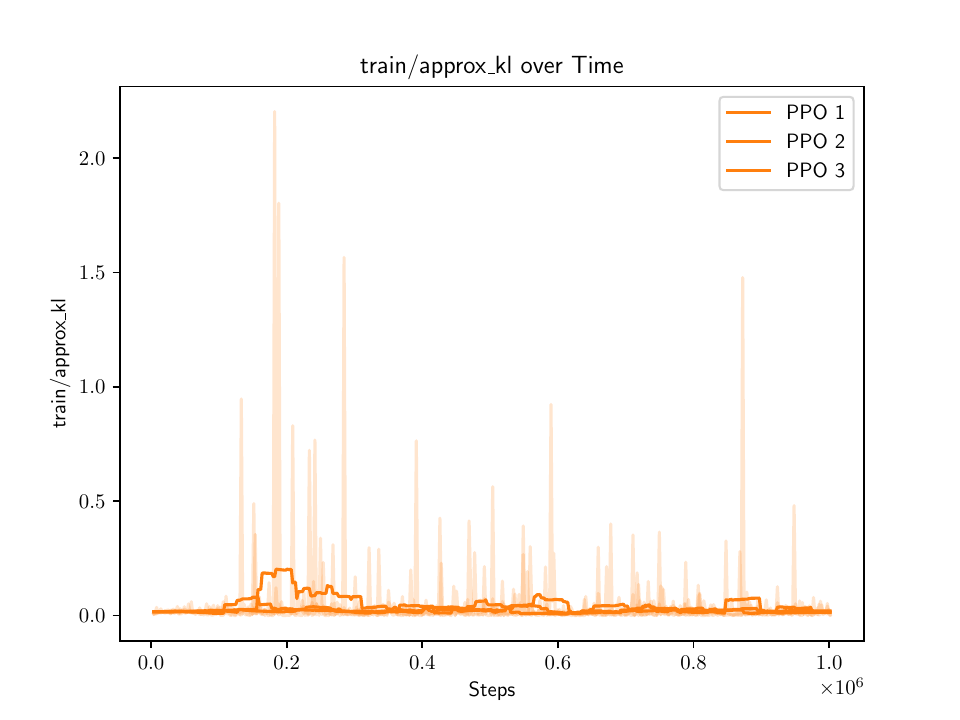}
\caption{Approximate KL divergence over time for PPO models.}
\label{fig:train_approx_kl}
\end{figure}

\begin{figure}[H]
\centering
\includegraphics[width=0.7\textwidth]{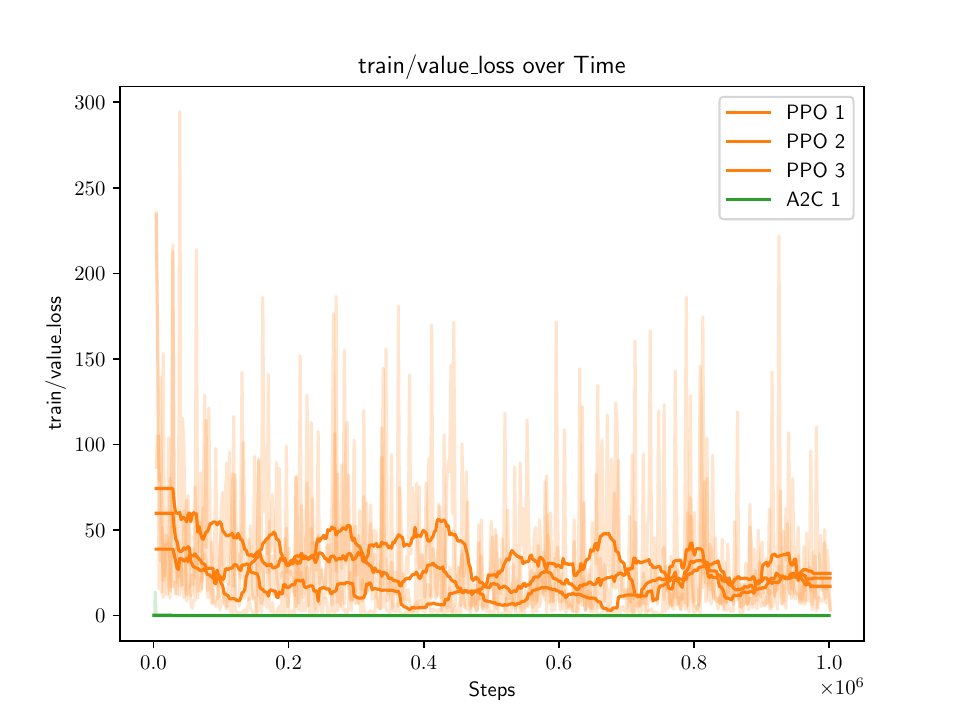}
\caption{Value loss over time for the PPO and A2C models.}
\label{fig:train_value_loss}
\end{figure}

\begin{figure}[H]
\centering
\includegraphics[width=0.7\textwidth]{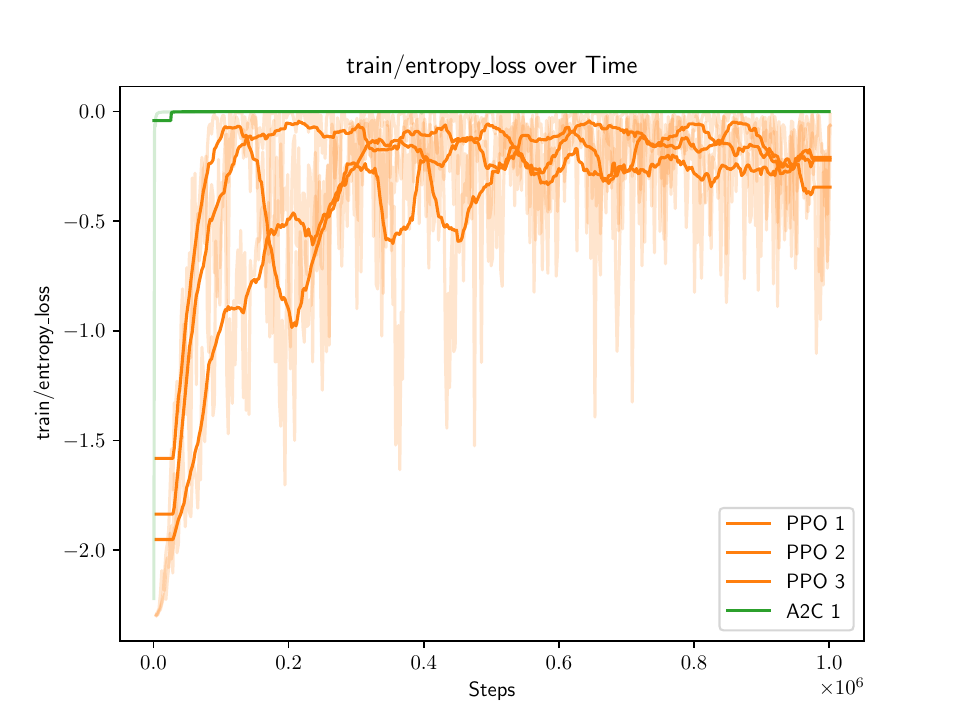}
\caption{Entropy loss over time, indicating the exploration behavior of the models.}
\label{fig:train_entropy_loss}
\end{figure}

\begin{figure}[H]
\centering
\includegraphics[width=0.7\textwidth]{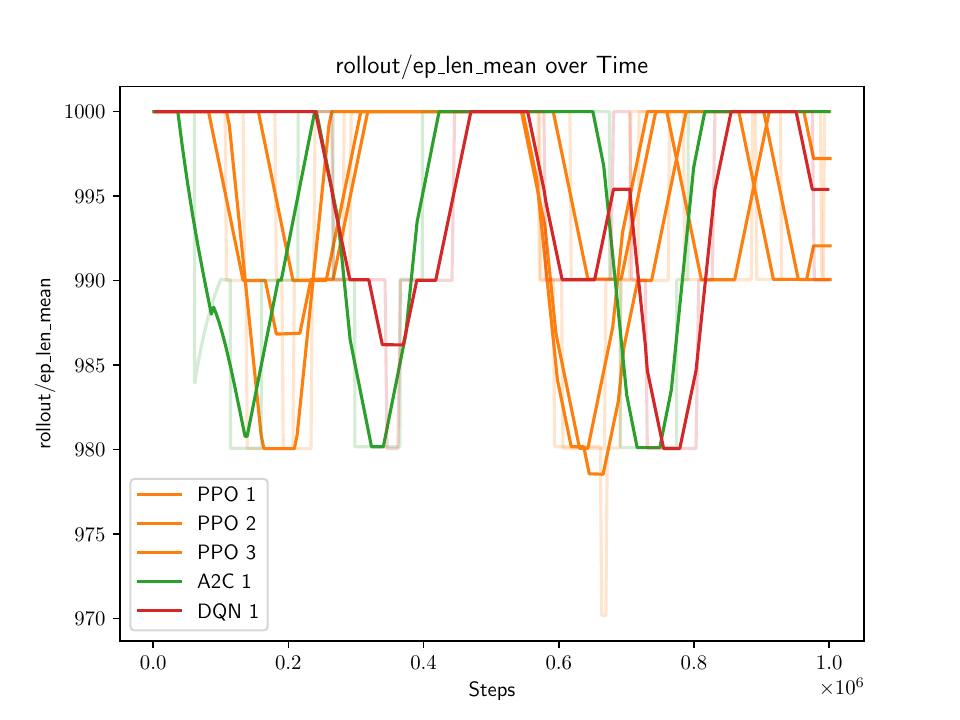}
\caption{Mean episode length over time, reflecting the consistency of the agents' performance.}
\label{fig:rollout_ep_len_mean}
\end{figure}

\begin{figure}[H]
\centering
\includegraphics[width=0.7\textwidth]{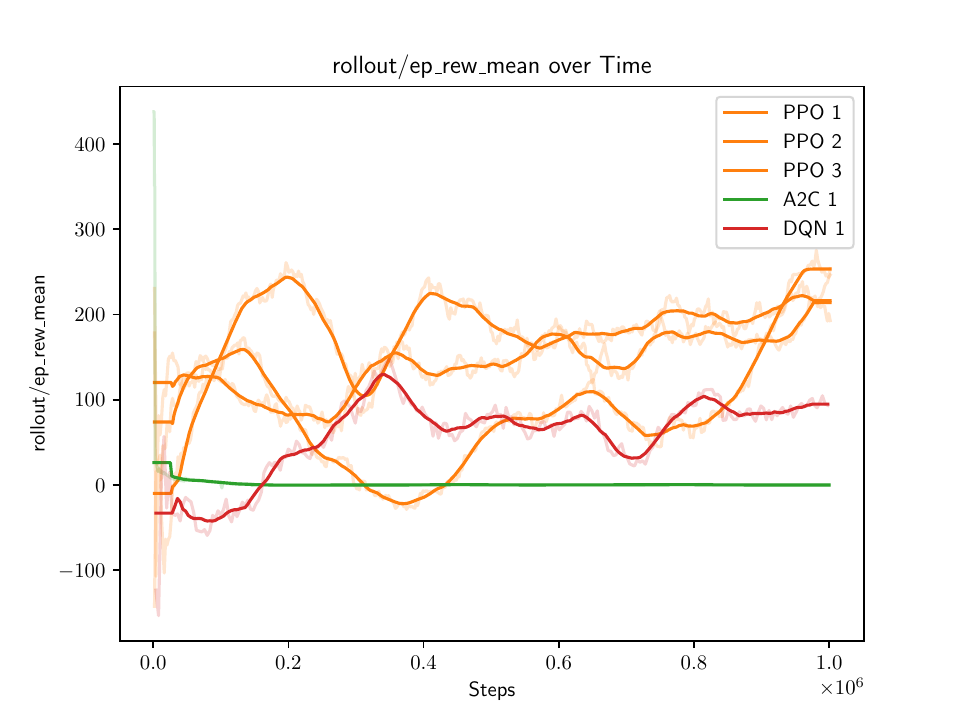}
\caption{Mean episode reward over time, an indicator of the agents' ability to maximize rewards.}
\label{fig:rollout_ep_rew_mean}
\end{figure}

\begin{figure}[H]
\centering
\includegraphics[width=0.7\textwidth]{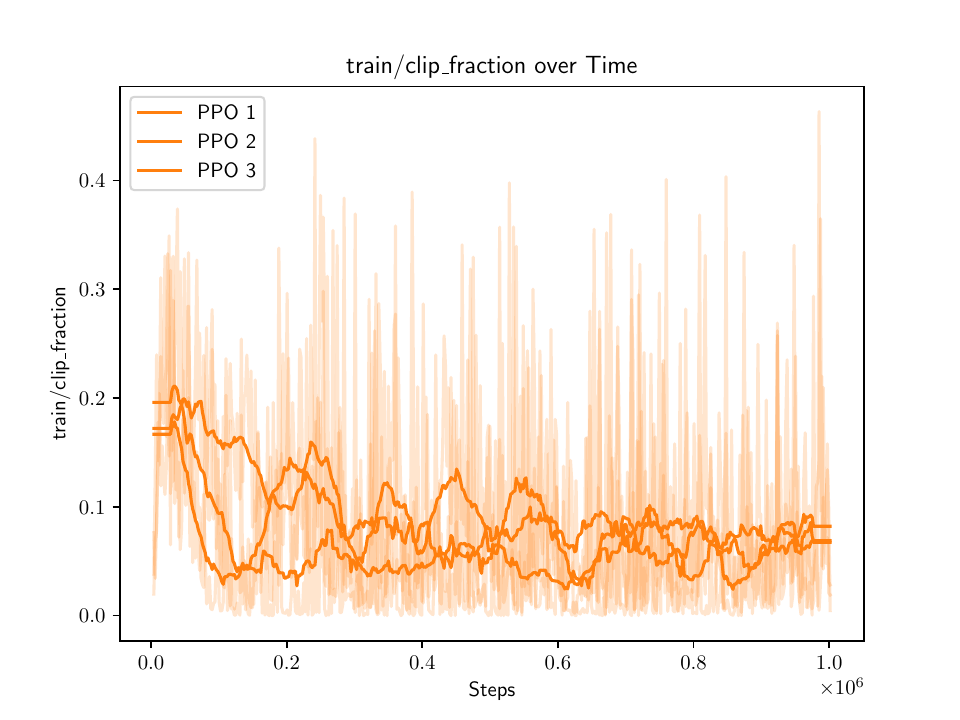}
\caption{Clip fraction over time for the PPO models, showing the proportion of clipped gradients during training.}
\label{fig:train_clip_fraction}
\end{figure}

\begin{figure}[H]
\centering
\includegraphics[width=0.7\textwidth]{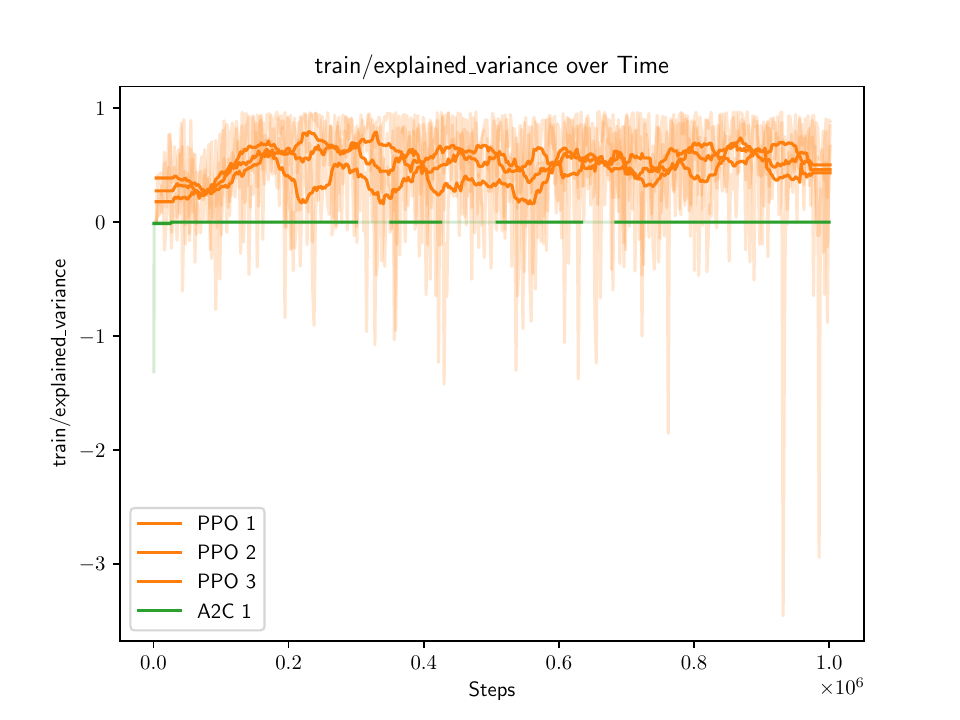}
\caption{Explained variance over time, indicating the predictive accuracy of the value function.}
\label{fig:train_explained_variance}
\end{figure}

\begin{figure}[H]
\centering
\includegraphics[width=0.7\textwidth]{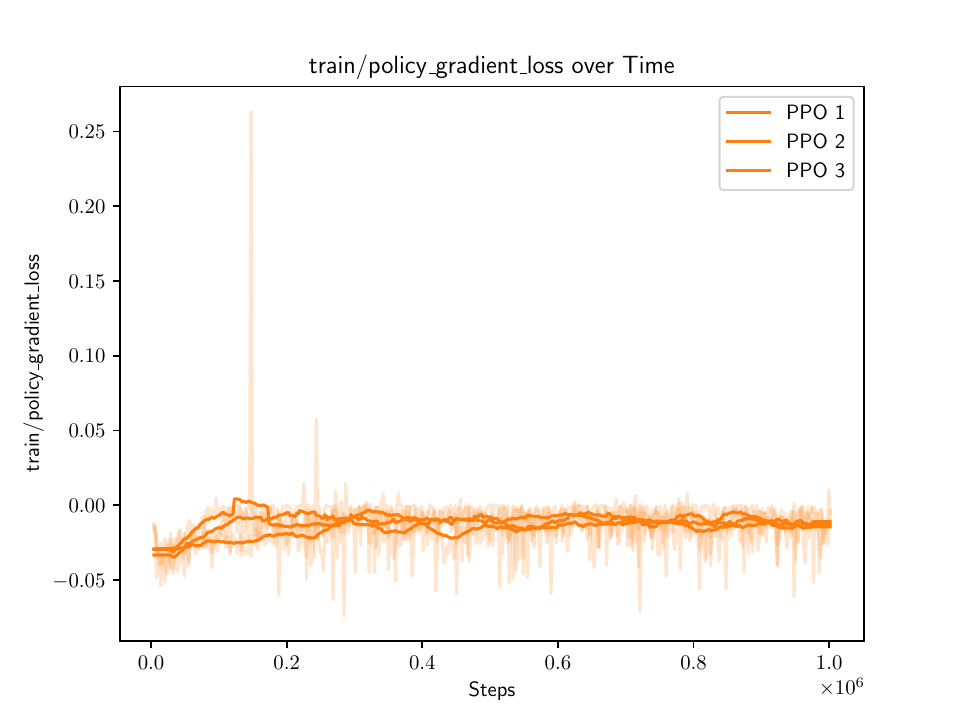}
\caption{Policy gradient loss over time, showing the optimization trajectory of the policy.}
\label{fig:train_policy_gradient_loss}
\end{figure}

Figure \ref{fig:time_fps} illustrates the FPS for the models. PPO models show a sharp decrease initially, which might suggest a computational adjustment period. Following this, all models appear to achieve a stable FPS, which is indicative of consistent computational load during training. Loss metrics, as shown in Figure \ref{fig:train_loss}, are crucial for understanding the learning progress. The PPO models display a general downward trend in loss, punctuated by spikes that suggest episodic learning or exploration phases. DQN's loss also decreases but shows more variability, which may indicate a different learning dynamic or exploration strategy. KL divergence is a measure of policy change. In Figure \ref{fig:train_approx_kl}, the PPO models demonstrate fluctuating KL divergence, with peaks corresponding to significant policy updates. The overall decline indicates stabilization of policy updates over time. Value loss, depicted in Figure \ref{fig:train_value_loss}, follows a similar pattern to the loss metric, with PPO showing a decrease over time, interspersed with spikes. A2C shows a less smooth trend, which could reflect a different approach to value estimation. Entropy loss, presented in Figure \ref{fig:train_entropy_loss}, measures the randomness in the policy. The declining trend seen with the PPO models suggests a reduction in policy entropy, which typically indicates increased confidence in action selection over time. Figure \ref{fig:rollout_ep_len_mean} shows the mean episode length. Sharp drops, particularly with the PPO and A2C models, suggest episodes where the agent may fail quickly, possibly due to exploring suboptimal policies. The mean episode reward, shown in Figure \ref{fig:rollout_ep_rew_mean}, is a direct indicator of agent performance. The general upward trend across models, with some variability, indicates learning and improvement in maximizing rewards. 
Clip fraction, as seen in Figure \ref{fig:train_clip_fraction}, is specific to PPO and indicates the fraction of times the clipping mechanism in the loss function is active. The general trend is downward, suggesting the policy is moving towards a stable region where clipping is less necessary. Explained variance, illustrated in Figure \ref{fig:train_explained_variance}, shows how well the value function predicts the actual returns. The values hovering around zero for A2C indicate poor prediction, whereas PPO shows some periods of better prediction. Finally, policy gradient loss, in Figure \ref{fig:train_policy_gradient_loss}, measures the optimization progress of the policy. PPO models display a stable trend with minor fluctuations, which suggests a steady optimization process.  In summary The analysis of the training performance metrics indicates that PPO models, on average, stabilize in terms of computational load, learning progress, and policy development. A2C and DQN exhibit different patterns, which may reflect their unique approaches to learning and exploration.

\section{MARL Results}

\subsection{Environment Description}

The \texttt{FireExtinguishingEnvParallel} class is a multi-agent reinforcement learning environment that simulates a scenario where agents (represented as bulldozers) work collaboratively to extinguish fires in a forest. This environment extends the \texttt{ParallelEnv} class from PettingZoo, indicating its suitability for parallel agents' interactions. Unlike single-agent environments, \texttt{FireExtinguishingEnvParallel} requires coordination and collaboration among multiple agents. Each agent operates independently, making decisions based on its observation space, but their actions collectively impact the overall environment and the task's success.

\subsubsection{Interaction with Frameworks}
Being a PettingZoo environment, it adheres to PettingZoo's API for multi-agent environments, enabling easy integration and interaction with other PettingZoo-compatible tools and libraries. SuperSuit can be used to wrap and transform this environment for advanced functionalities like vectorization, frame-stacking, etc., making it more flexible for different training setups. To integrate with SB3, a popular reinforcement learning library, the environment needs to be converted into a vectorized form using SuperSuit. This allows leveraging SB3's algorithms for training agents in the environment. While primarily based on PettingZoo's API, the environment maintains compatibility with Gym-like interfaces through wrappers, ensuring it can be used with Gym-based tools and libraries.
\texttt{FireExtinguishingEnvParallel} presents a complex and dynamic multi-agent reinforcement learning challenge, ideal for studying collaborative strategies and agent interactions in a shared environment.

\subsection{Multi-Agent and Model Training}

The training process begins with the initialization of the multi-agent environment. This environment simulates a scenario where multiple agents collaborate to extinguish fires in a forested grid. Parameters such as grid size, vision range, and the number of agents are defined at this stage.

Following initialization, the environment is wrapped with the PettingZoo API to standardize the multi-agent interface. This step ensures compatibility with further processing and vectorization tools. The environment is then vectorized and concatenated using SuperSuit, preparing it for integration with Stable Baselines3 (SB3), a reinforcement learning library.
Centralized training is conducted using SB3's Proximal Policy Optimization (PPO) algorithm. In this phase, a shared policy model is trained across all agents, enabling them to learn cooperative strategies. The model undergoes iterative training, where agent experiences (observations, actions, rewards) are collected and used to improve the policy.

During training, checkpoints and model states are saved periodically. This allows for the preservation of learning progress and facilitates model evaluation and deployment. Upon successful training, the model is deployed for decentralized execution. Each agent, following the shared policy, independently decides its actions based on its local observations. The environment responds to these actions, updating its state and providing new observations and rewards to the agents.

\begin{figure}[H]
\centering
\includegraphics[width=0.9\textwidth]{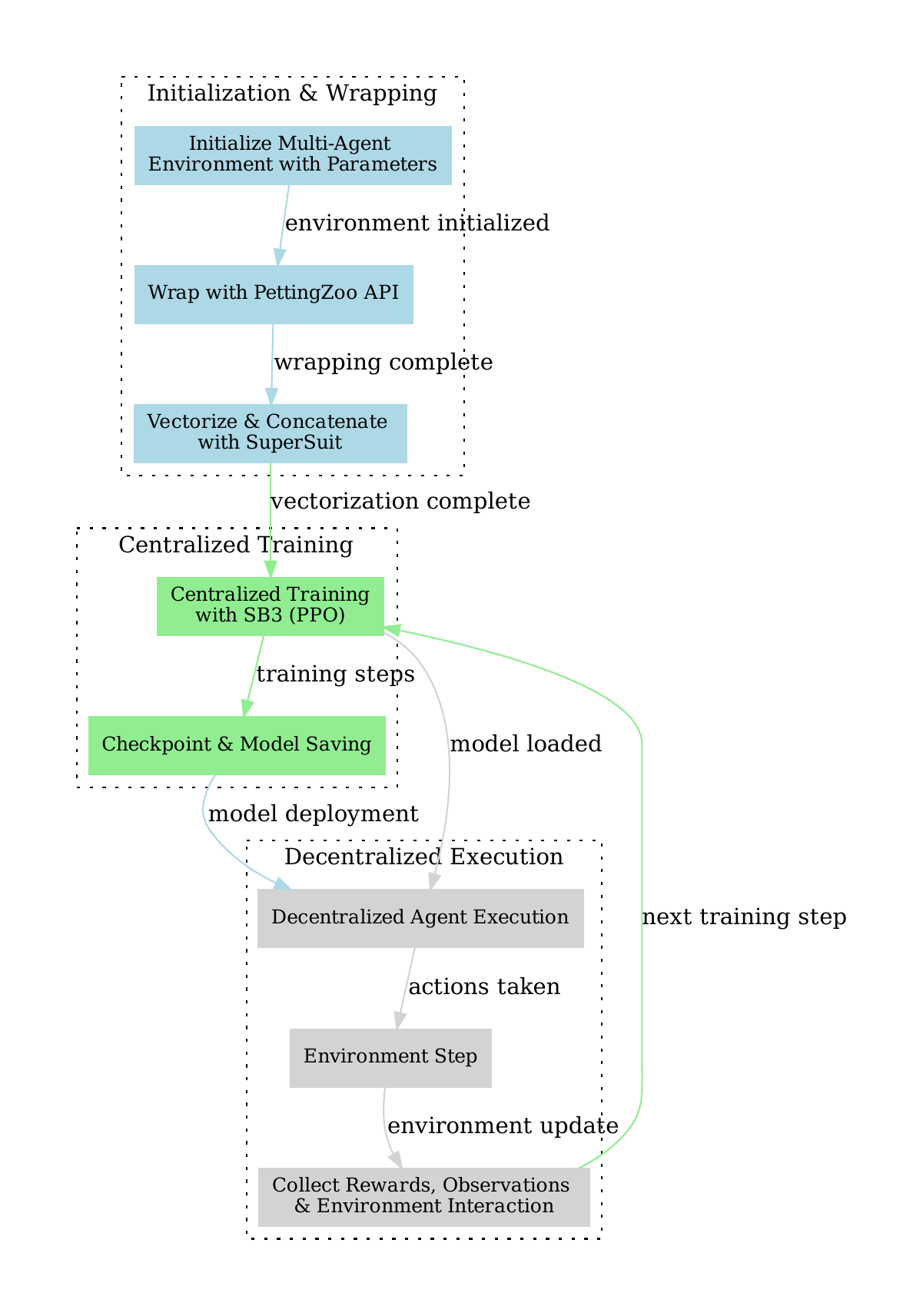}
\caption{Flowchart of the Multi-Agent Training Process in \textbf{FireExtinguishingEnvParallel}}
\label{fig:fire_ext_env_workflow_multi}

\end{figure}

This loop of action-taking and environment response continues until the termination conditions are met (e.g., all fires extinguished or maximum steps reached). The training and execution process in the \texttt{FireExtinguishingEnvParallel} environment demonstrates the efficiency of combining centralized training with decentralized execution in a multi-agent setting. This approach enables agents to learn collaborative behaviors while retaining the ability to act independently based on localized information. The training process can be shown in Figure \ref{fig:fire_ext_env_workflow_multi}, which intricately showcases how the Multi-agent training process is realized.

\section{Performance Metrics}

\section{Performance Metrics Discussion}

\begin{figure}[H]
\centering
\includegraphics[width=0.7\textwidth]{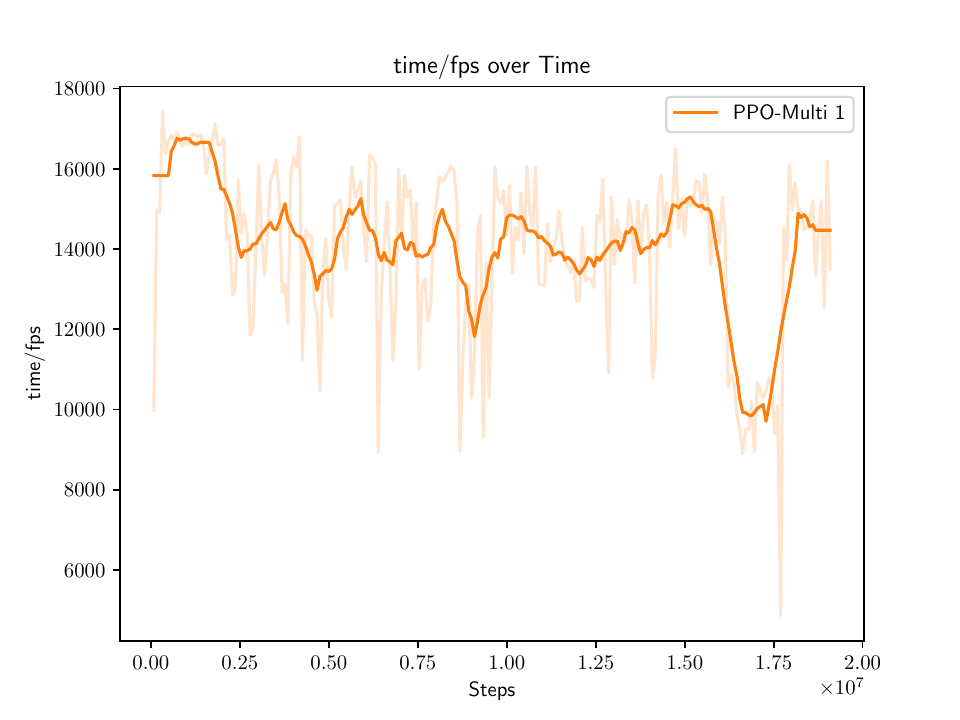}
\caption{Time per update/frame per second (FPS) over-training steps}
\label{fig:time_fps_Multi}
\end{figure}

\begin{figure}[H]
\centering
\includegraphics[width=0.7\textwidth]{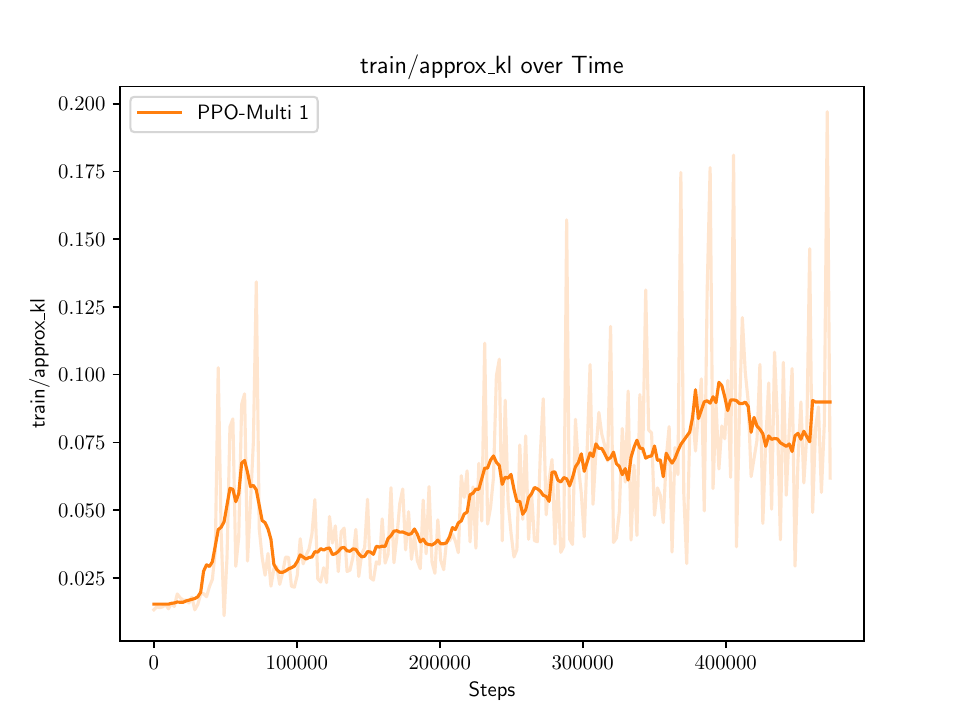}
\caption{Approximate Kullback-Leibler divergence over training steps}
\label{fig:approx_kl_Multi}
\end{figure}

\begin{figure}[H]
\centering
\includegraphics[width=0.7\textwidth]{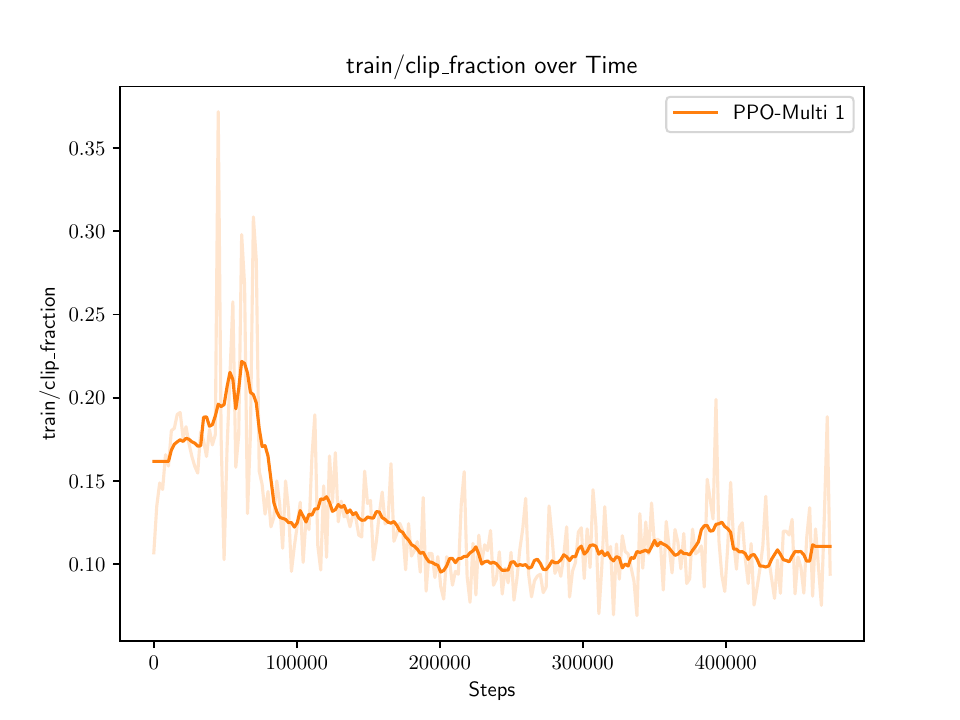}
\caption{Clip fraction over training steps}
\label{fig:clip_fraction_Multi}
\end{figure}

\begin{figure}[H]
\centering
\includegraphics[width=0.7\textwidth]{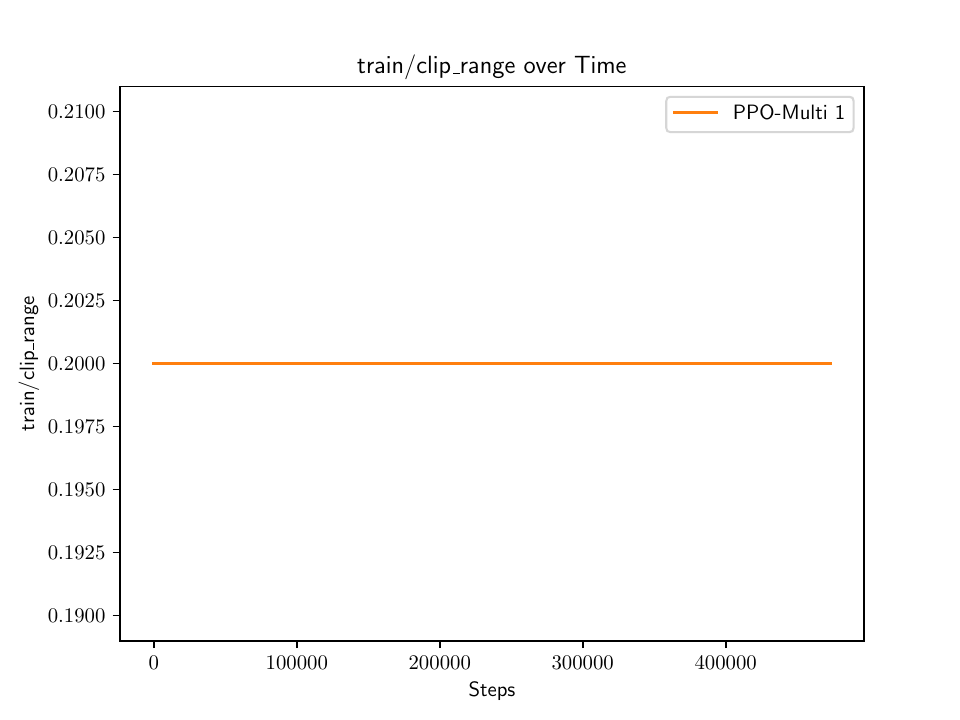}
\caption{Clip range over training steps}
\label{fig:clip_range_Multi}
\end{figure}

\begin{figure}[H]
\centering
\includegraphics[width=0.7\textwidth]{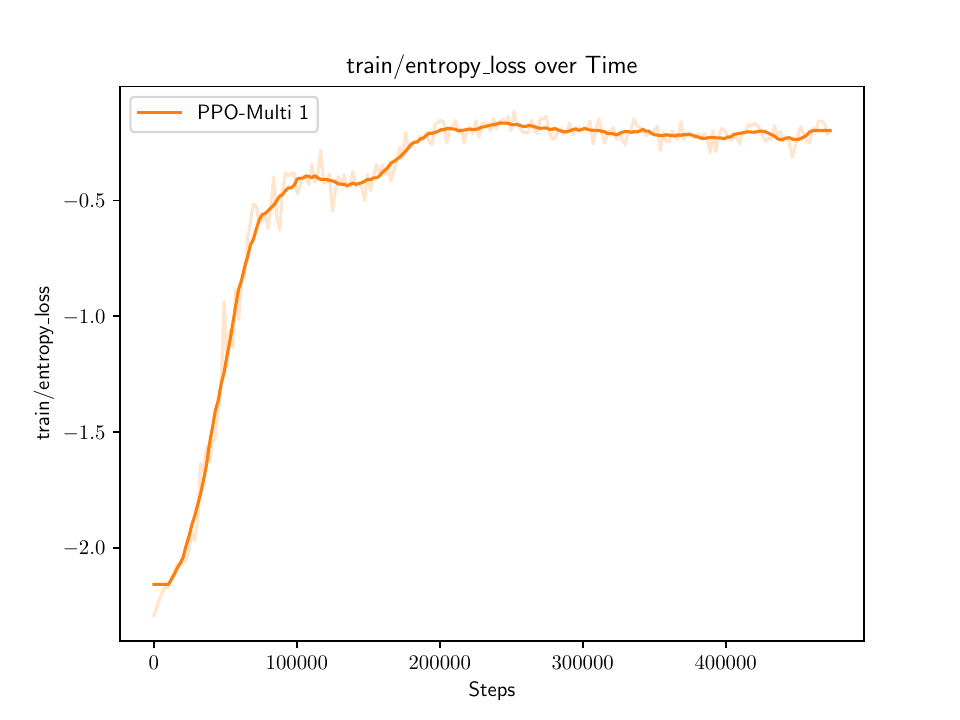}
\caption{Entropy loss over training steps}
\label{fig:entropy_loss_Multi}
\end{figure}

\begin{figure}[H]
\centering
\includegraphics[width=0.7\textwidth]{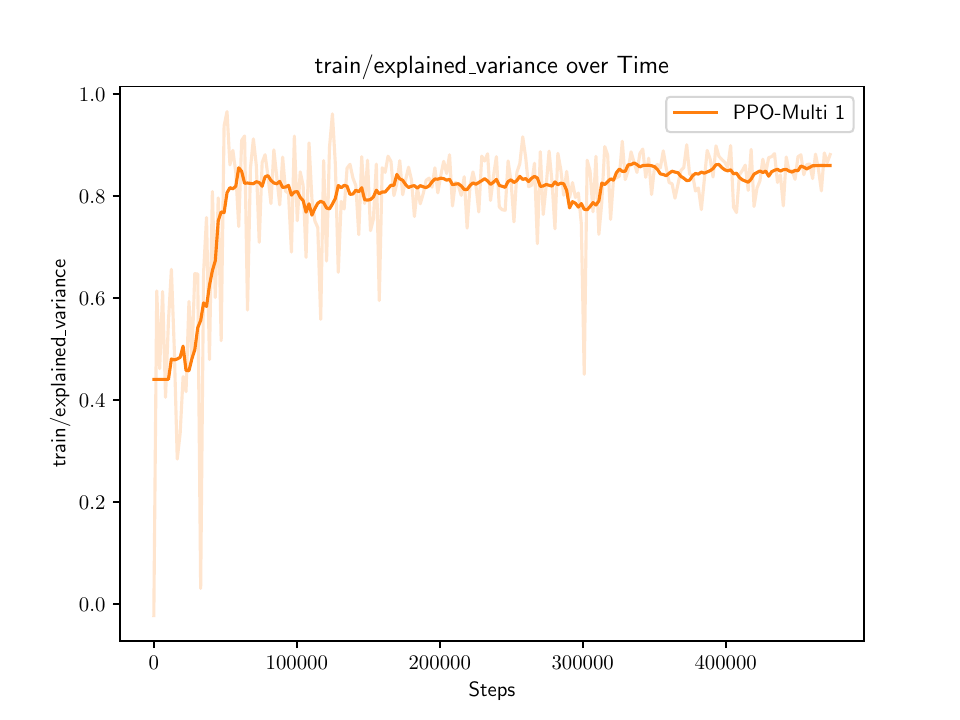}
\caption{Explained variance over training steps}
\label{fig:explained_variance_Multi}
\end{figure}

\begin{figure}[H]
\centering
\includegraphics[width=0.7\textwidth]{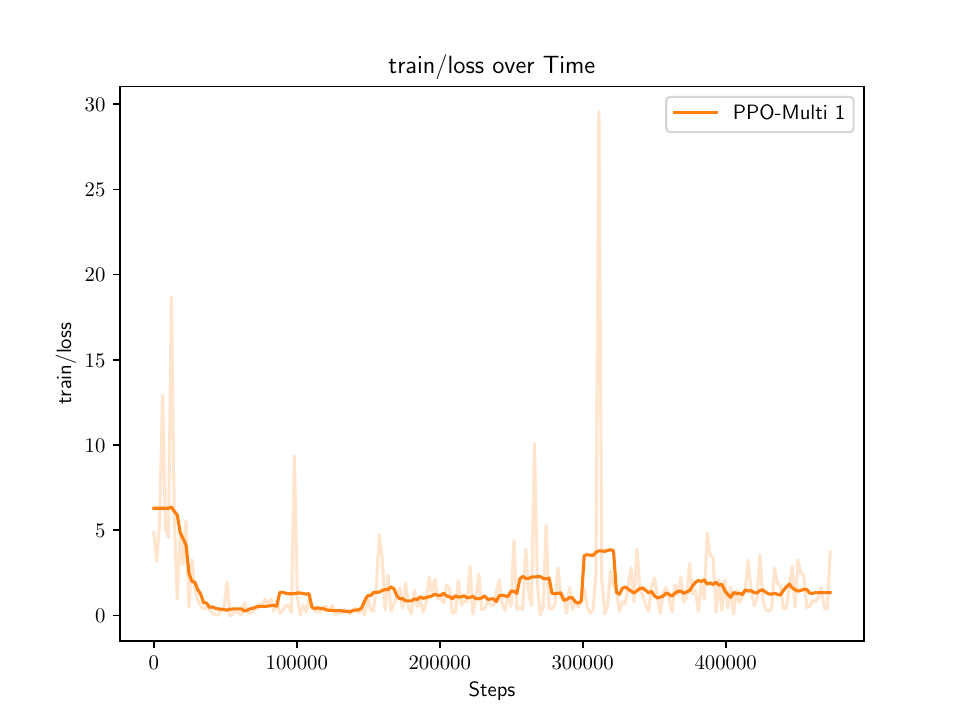}
\caption{Loss over training steps}
\label{fig:loss_Multi}
\end{figure}

\begin{figure}[H]
\centering
\includegraphics[width=0.7\textwidth]{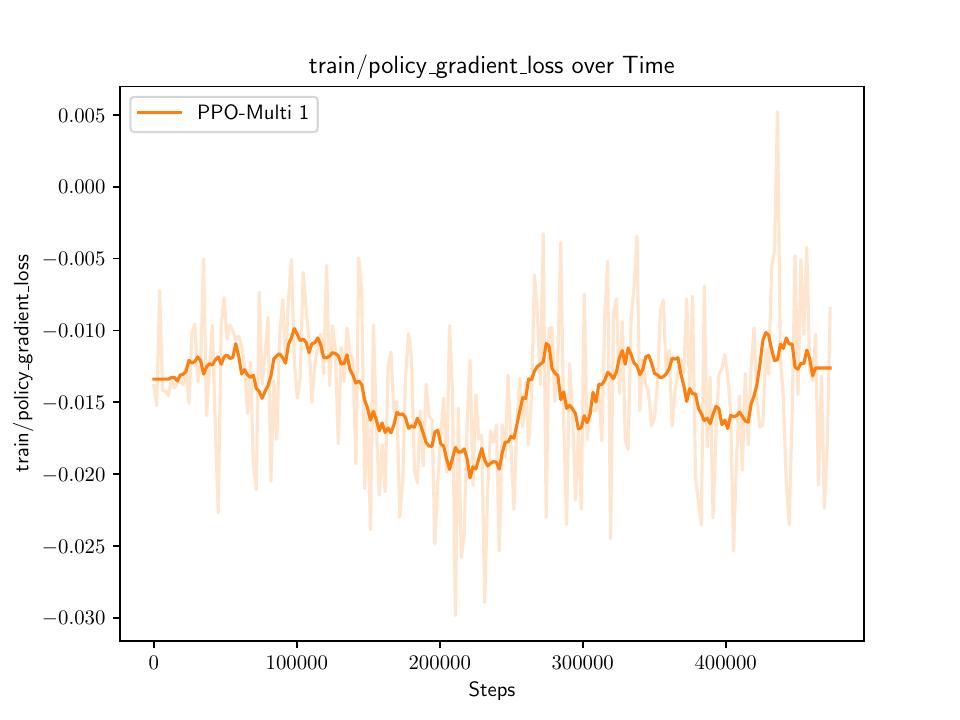}
\caption{Policy gradient loss over training steps}
\label{fig:policy_gradient_loss_Multi}
\end{figure}

\begin{figure}[H]
\centering
\includegraphics[width=0.7\textwidth]{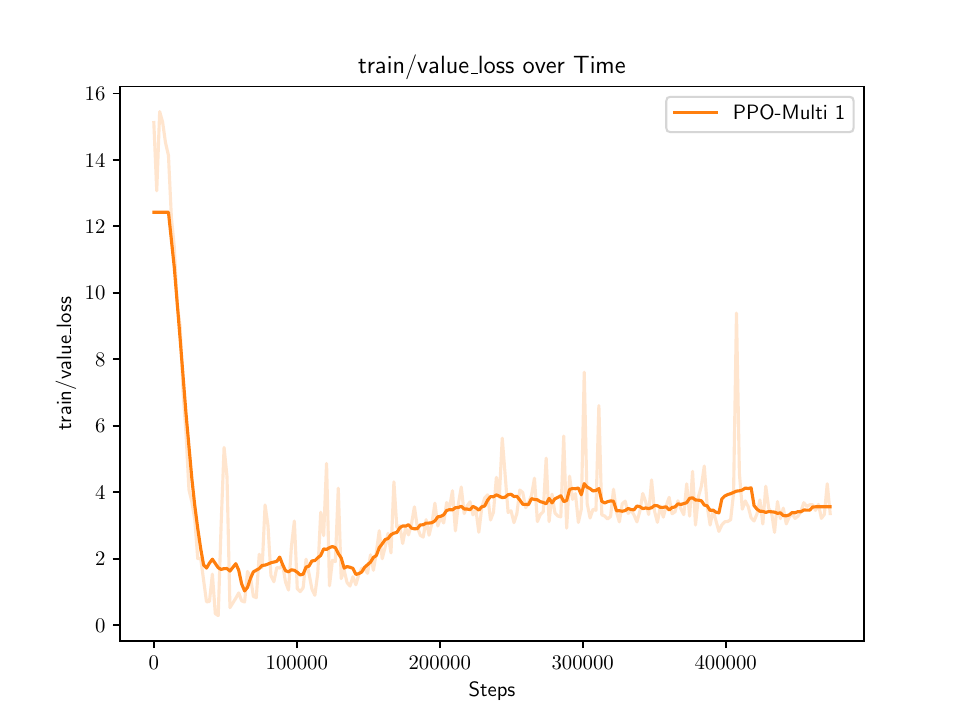}
\caption{Value loss over training steps}
\label{fig:value_loss_Multi}
\end{figure}

During the training of our multi-agent reinforcement learning framework using the PPO algorithm, we observed several key performance metrics which are vital for understanding the learning dynamics and effectiveness of the model. Figures \ref{fig:time_fps_Multi}, \ref{fig:approx_kl_Multi}, \ref{fig:clip_fraction_Multi}, \ref{fig:clip_range_Multi}, \ref{fig:entropy_loss_Multi}, \ref{fig:explained_variance_Multi}, \ref{fig:loss_Multi}, \ref{fig:policy_gradient_loss_Multi}, and \ref{fig:value_loss_Multi} showcase these metrics over the course of training steps.

Figure \ref{fig:time_fps_Multi} shows the time and frames per second (FPS) over the training steps. The FPS provides insight into the computational efficiency of the training process. An overall downward trend or significant drops in FPS may indicate computational bottlenecks or increased complexity in the simulation as the training progresses. The Approximate KL Divergence, depicted in Figure \ref{fig:approx_kl_Multi}, measures how the policy distribution changes over time. Spikes in KL divergence can suggest significant policy updates, which could either be beneficial as the agents explore new strategies or detrimental if the policy diverges too much from the previous one, potentially leading to instability in training.
The Clip Fraction, shown in Figure \ref{fig:clip_fraction_Multi}, indicates the fraction of the agent's probability ratio that was clipped by the PPO algorithm. This metric helps in understanding the extent to which the policy is being constrained and can also reflect the stability of the training process. Figure \ref{fig:clip_range_Multi} shows the clip range, which is the range within which the probability ratio is constrained. A constant clip range, as observed, suggests a stable constraint over the policy updates throughout the training. Entropy loss, visualized in Figure \ref{fig:entropy_loss_Multi}, describes the randomness in the policy distribution. A higher entropy can encourage exploration by the agents, while a lower entropy indicates a more deterministic policy. Ideally, entropy should decrease over time as the policy converges to an optimal strategy. The explained variance metric, in Figure \ref{fig:explained_variance_Multi}, measures how well the value function predicts the rewards. Values closer to 1 indicate better predictions, which can lead to more efficient learning by the agents. Figures \ref{fig:loss_Multi}, \ref{fig:policy_gradient_loss_Multi}, and \ref{fig:value_loss_Multi} provide insights into the loss experienced by the policy and value functions. The loss metrics are critical in monitoring the convergence of the learning process. Sharp increases could indicate potential issues that need to be addressed for smoother learning.

\chapter{Conclusion}
In this project, we have explored the application of single and multi-agent reinforcement learning frameworks to the challenging problem of wildfire management. The sophisticated simulation environment we developed integrates CA for fire spread and Perlin noise for terrain generation. This environment not only captures the complexity of wildfire dynamics but also provides a fertile ground for RL agents to learn and optimize strategies for fire extinguishing and forest preservation.

\subsection*{Key Findings}
The experimentation phase yielded several key insights. Firstly, we found that the Proximal Policy Optimization (PPO) algorithm adapted well to the intricacies of the fire management scenario, achieving a balance between exploration and exploitation, as evidenced by the performance metrics discussed in previous chapters. The training process demonstrated a convergence of policy, which is indicative of the agents' increasing proficiency in tackling the wildfire management task. Our MARL framework showed promise in fostering collaborative behaviors among agents. The decentralized execution post-training allowed agents to operate independently yet effectively, considering the global objectives of minimizing fire damage and preserving the forest, adhering to CTDE training conditions of independent agents.

Despite these successes, we encountered several challenges. Computational efficiency remained a concern, particularly in the MARL setting, where the increase in agents led to a more complex state space and action space. Moreover, the initial instability during training, reflected by the fluctuations in the frames per second (FPS), suggested that our computational resources were stretched at times.

Additionally, while the PPO algorithm was effective, it was not without shortcomings. The occasional spikes in the KL divergence and loss metrics hinted at moments where the agents' policies could potentially deviate from optimal behaviors. Furthermore, the fixed parameters of Perlin noise and the simplified wind model, although computationally efficient, might have limited the agents' exposure to more varied and realistic fire scenarios.

\subsection*{Final Thoughts}

The results of our experimentation have demonstrated the potential of applying reinforcement learning to the domain of wildfire management. The project has laid a foundation for future research to build upon, with the ultimate goal of developing AI agents capable of assisting in the critical task of managing and mitigating wildfires, a growing concern in the face of climate change.

The journey from algorithmic inception to practical application is often fraught with unexpected challenges and learning opportunities. In our case, while we achieved a significant measure of success, the path ahead is clear—there is much room for improvement, refinement, and, most importantly, innovation.

\subsection*{Future Directions}

Looking forward, there are several avenues to enhance our wildfire management simulation and learning frameworks. Investigating more complex and dynamic models for wind and weather patterns could offer a more realistic challenge to the agents. Advanced MARL algorithms that can efficiently handle the increased complexity of such models would be worth exploring.

In addition, adapting the learning algorithms to leverage parallel computational resources more effectively could address the computational bottlenecks observed. The inclusion of human-in-the-loop training could also provide a unique perspective, allowing the agents to learn from human expertise in wildfire management.
\subsection*{Introducing Conservation Metric}

\subsubsection*{Reward Shaping (Conservation Metric)}
\paragraph*{Objective}
The reward system is designed to incentivize agents to preserve as many trees as possible, efficiently extinguish fires, and collaboratively work without causing unnecessary environmental damage. The system aims to balance the actions taken against fires with the imperative to conserve the forest.

\paragraph*{Definitions}
Let's define the variables used in the reward functions:

\begin{itemize}
    \item \( T_p \): Number of trees preserved (untouched by fire or agents).
    \item \( T_e \): Number of trees successfully extinguished (saved from burning).
    \item \( T_c \): Number of trees cleared (by agents, potentially to create firebreaks).
    \item \( T_b \): Number of trees burned (lost to fire).
    \item \( A_{total} \): Total number of actions taken by all agents.
    \item \( T_{total} \): Total time taken to control the fires.
    \item \( r_{protect, area} \): Reward for protecting each unit area from fire.
    \item \( \alpha, \beta, \gamma, \delta, \epsilon, \zeta, \eta, \theta_{i} \): Coefficients representing the weight of each component in the reward calculation.
\end{itemize}

\paragraph*{Reward Functions}

\textbf{1. Efficiency-Oriented Reward Function}
This function rewards agents for preserving trees and extinguishing fires while penalizing unnecessary clearing and trees burned. It also considers the total number of actions taken to encourage efficient decision-making.

\[ R = \alpha \cdot T_p + \beta \cdot T_e - \gamma \cdot T_c - \delta \cdot T_b - \epsilon \cdot A_{total} \]

\textbf{2. Time-Based Reward Function}
This function emphasizes the importance of time in managing fires. It rewards quick and effective action while penalizing the total time taken and trees lost.

\[ R = \alpha \cdot T_p + \beta \cdot T_e - \gamma \cdot T_c - \delta \cdot T_b - \zeta / T_{total} \]

\textbf{3. Spatial Coverage Reward Function}
This function rewards agents for protecting large areas of the forest from fire, in addition to the other components of preserving, extinguishing, clearing, and burned trees.

\[ R = \alpha \cdot T_p + \beta \cdot T_e - \gamma \cdot T_c - \delta \cdot T_b + \sum_{area} r_{protect, area} \]

\textbf{4. Collaborative Reward Function}
This function promotes teamwork by scaling rewards based on collaborative efforts, encouraging agents to work together efficiently.

\[ R = \eta \cdot (\alpha \cdot T_p + \beta \cdot T_e - \gamma \cdot T_c - \delta \cdot T_b) \]

\textbf{5. Dynamic Adaptive Reward Function}
This function dynamically adjusts the rewards based on specific environmental states or scenarios, encouraging adaptive behavior.

\[ R = \alpha \cdot T_p + \beta \cdot T_e - \gamma \cdot T_c - \delta \cdot T_b + \sum_{i=1}^{N} \theta_{i} \cdot r_{dynamic, i} \]

\paragraph{Purpose of the Design}
The reward design's purpose is to create a balanced and holistic approach to managing wildfires in a simulated environment. It encourages preservation, efficient action, and collaboration, while also allowing for adaptive responses to dynamic situations.


\bibliographystyle{IEEEtran}
\bibliography{ref}

\begin{thebibliography}{10}
\providecommand{\url}[1]{#1}
\csname url@samestyle\endcsname
\providecommand{\newblock}{\relax}
\providecommand{\bibinfo}[2]{#2}
\providecommand{\BIBentrySTDinterwordspacing}{\spaceskip=0pt\relax}
\providecommand{\BIBentryALTinterwordstretchfactor}{4}
\providecommand{\BIBentryALTinterwordspacing}{\spaceskip=\fontdimen2\font plus
\BIBentryALTinterwordstretchfactor\fontdimen3\font minus
  \fontdimen4\font\relax}
\providecommand{\BIBforeignlanguage}[2]{{%
\expandafter\ifx\csname l@#1\endcsname\relax
\typeout{** WARNING: IEEEtran.bst: No hyphenation pattern has been}%
\typeout{** loaded for the language `#1'. Using the pattern for}%
\typeout{** the default language instead.}%
\else
\language=\csname l@#1\endcsname
\fi
#2}}
\providecommand{\BIBdecl}{\relax}
\BIBdecl

\bibitem{tymstra_wildfire_2020}
\BIBentryALTinterwordspacing
C.~Tymstra, B.~J. Stocks, X.~Cai, and M.~D. Flannigan, ``Wildfire management in
  {Canada}: {Review}, challenges and opportunities,'' \emph{Progress in
  Disaster Science}, vol.~5, p. 100045, Jan. 2020. [Online]. Available:
  \url{https://www.sciencedirect.com/science/article/pii/S2590061719300456}
\BIBentrySTDinterwordspacing

\bibitem{noauthor_canadian_2023}
\BIBentryALTinterwordspacing
``\BIBforeignlanguage{en-US}{The “{Canadian}” {Wildfires} of 2023},'' Jun.
  2023. [Online]. Available:
  \url{https://carleton.ca/thedisasterlab/2023/the-canadian-wildfires-of-2023/}
\BIBentrySTDinterwordspacing

\bibitem{noauthor_international_nodate}
\BIBentryALTinterwordspacing
``\BIBforeignlanguage{en-US}{International {Journal} of {Wildland} {Fire}
  ({IJWF})}.'' [Online]. Available:
  \url{https://www.iawfonline.org/international-journal-wildland-fire-ijwf/}
\BIBentrySTDinterwordspacing

\bibitem{news_wildfires_nodate}
\BIBentryALTinterwordspacing
A.~B.~C. News, ``\BIBforeignlanguage{en}{Wildfires in {Canada} have broken
  records for area burned, evacuations and cost, official says}.'' [Online].
  Available:
  \url{https://abcnews.go.com/International/wireStory/wildfires-canada-broken-records-area-burned-evacuations-cost-100806230}
\BIBentrySTDinterwordspacing

\bibitem{pereira_review_2022}
\BIBentryALTinterwordspacing
J.~Pereira, J.~Mendes, J.~S.~S. Júnior, C.~Viegas, and J.~R. Paulo,
  ``\BIBforeignlanguage{en}{A {Review} of {Genetic} {Algorithm} {Approaches}
  for {Wildfire} {Spread} {Prediction} {Calibration}},''
  \emph{\BIBforeignlanguage{en}{Mathematics}}, vol.~10, no.~3, p. 300, Jan.
  2022, number: 3 Publisher: Multidisciplinary Digital Publishing Institute.
  [Online]. Available: \url{https://www.mdpi.com/2227-7390/10/3/300}
\BIBentrySTDinterwordspacing

\bibitem{perestrelo_modelling_2021}
\BIBentryALTinterwordspacing
S.~Perestrelo, M.~C. Grácio, N.~A. Ribeiro, and L.~M. Lopes,
  ``\BIBforeignlanguage{en}{Modelling {Forest} {Fires} {Using} {Complex}
  {Networks}},'' \emph{\BIBforeignlanguage{en}{Mathematical and Computational
  Applications}}, vol.~26, no.~4, p.~68, Dec. 2021, number: 4 Publisher:
  Multidisciplinary Digital Publishing Institute. [Online]. Available:
  \url{https://www.mdpi.com/2297-8747/26/4/68}
\BIBentrySTDinterwordspacing

\bibitem{xu2022modeling}
Y.~Xu, D.~Li, H.~Ma, R.~Lin, and F.~Zhang, ``Modeling forest fire spread using
  machine learning-based cellular automata in a gis environment,''
  \emph{Forests}, vol.~13, no.~12, p. 1974, 2022.

\bibitem{byari_multi-scale_2022}
\BIBentryALTinterwordspacing
M.~Byari, A.~Bernoussi, O.~Jellouli, M.~Ouardouz, and M.~Amharref,
  ``Multi-{Scale} 3d {Cellular} {Automata} {Modeling}: {Application} to
  {Wildland} {Fire} {Spread},'' \emph{SSRN Electronic Journal}, vol. null, p.
  null, 2022. [Online]. Available:
  \url{https://www.semanticscholar.org/paper/6716d5a3790b3825c3fba326bf017103f1092787}
\BIBentrySTDinterwordspacing

\bibitem{liu2019emergent}
S.~Liu, G.~Lever, J.~Merel, S.~Tunyasuvunakool, N.~Heess, and T.~Graepel,
  ``Emergent coordination through competition,'' \emph{International Conference
  on Learning Representations}, 2019.

\bibitem{yin_simulation_2018}
\BIBentryALTinterwordspacing
H.~Yin, H.~Jin, Y.~Zhao, Y.-L. Fan, L.~Qin, Q.~Chen, L.~Huang, X.~Jia, L.~Liu,
  Y.~Dai, and Y.~Xiao, ``The simulation of surface fire spread based on
  {Rothermel} model in windthrow area of {Changbai} {Mountain} ({Jilin},
  {China}),'' 2018. [Online]. Available:
  \url{https://www.semanticscholar.org/paper/1e215cc03da2044d252b7107f91d1bcc36a84d74}
\BIBentrySTDinterwordspacing

\bibitem{zhang_study_2021}
\BIBentryALTinterwordspacing
S.~Zhang, J.~Liu, H.~Gao, X.~Chen, X.~Li, and J.~Hua, ``Study on {Forest}
  {Fire} spread {Model} of {Multi}-dimensional {Cellular} {Automata} based on
  {Rothermel} {Speed} {Formula},'' \emph{CERNE}, vol. null, p. null, 2021.
  [Online]. Available:
  \url{https://www.semanticscholar.org/paper/e9f829f0187edaa09259e1233881fed0b55d9d29}
\BIBentrySTDinterwordspacing

\bibitem{wahyuni_investigating_2021}
\BIBentryALTinterwordspacing
I.~Wahyuni, A.~Shabrina, and A.~Latifah, ``Investigating {Multivariable}
  {Factors} of the {Southern} {Borneo} {Forest} and {Land} {Fire} based on
  {Random} {Forest} {Model},'' \emph{Proceedings of the 2021 International
  Conference on Computer, Control, Informatics and Its Applications}, vol.
  null, p. null, 2021. [Online]. Available:
  \url{https://www.semanticscholar.org/paper/c48c80cf0327f9d0ac55b4a3bc77090aa155b7ce}
\BIBentrySTDinterwordspacing

\bibitem{fischer_autobiomes_2020}
\BIBentryALTinterwordspacing
R.~Fischer, P.~Dittmann, R.~Weller, and G.~Zachmann,
  ``\BIBforeignlanguage{en}{{AutoBiomes}: procedural generation of multi-biome
  landscapes},'' \emph{\BIBforeignlanguage{en}{The Visual Computer}}, vol.~36,
  no.~10, pp. 2263--2272, Oct. 2020. [Online]. Available:
  \url{https://doi.org/10.1007/s00371-020-01920-7}
\BIBentrySTDinterwordspacing

\bibitem{mastorakos_hybrid_2022}
\BIBentryALTinterwordspacing
E.~Mastorakos, S.~Gkantonas, G.~Efstathiou, and A.~Giusti, ``A hybrid
  stochastic {Lagrangian} – cellular automata framework for modelling fire
  propagation in inhomogeneous terrains,'' \emph{Proceedings of the Combustion
  Institute}, vol. null, p. null, 2022. [Online]. Available:
  \url{https://www.semanticscholar.org/paper/8df63a09953ca72e07077f1b533e18278f7af182}
\BIBentrySTDinterwordspacing

\bibitem{le_procedural_nodate}
T.~Le, ``\BIBforeignlanguage{English}{Procedural {Terrain} {Generation} {Using}
  {Perlin} {Noise}},'' publisher: California State Polytechnic University,
  Pomona.

\bibitem{jain_adaptive_2023}
\BIBentryALTinterwordspacing
A.~Jain, A.~Sharma, and Rajan, ``Adaptive \& {Multi}-{Resolution} {Procedural}
  {Infinite} {Terrain} {Generation} with {Diffusion} {Models} and {Perlin}
  {Noise},'' in \emph{Proceedings of the {Thirteenth} {Indian} {Conference} on
  {Computer} {Vision}, {Graphics} and {Image} {Processing}}, ser. {ICVGIP}
  '22.\hskip 1em plus 0.5em minus 0.4em\relax New York, NY, USA: Association
  for Computing Machinery, May 2023, pp. 1--9. [Online]. Available:
  \url{https://doi.org/10.1145/3571600.3571657}
\BIBentrySTDinterwordspacing

\bibitem{rosadi_prediction_2021}
\BIBentryALTinterwordspacing
D.~Rosadi, W.~Andriyani, D.~Arisanty, and D.~Agustina, ``Prediction of {Forest}
  {Fire} using {Hybrid} {SOM}-{AdaBoost} {Method},'' \emph{Journal of Physics:
  Conference Series}, vol. 2123, p. null, 2021. [Online]. Available:
  \url{https://www.semanticscholar.org/paper/788662cc8bb97848476ab0448a15d5f90ba64398}
\BIBentrySTDinterwordspacing

\bibitem{zhao_simulation_2021}
\BIBentryALTinterwordspacing
Y.~Zhao and D.~Geng, ``Simulation of {Forest} {Fire} {Occurrence} and {Spread}
  {Based} on {Cellular} {Automata} {Model},'' \emph{2021 2nd International
  Conference on Artificial Intelligence and Information Systems}, vol. null, p.
  null, 2021. [Online]. Available:
  \url{https://www.semanticscholar.org/paper/01a2c3f6864d9ecd8bd05b0bd3e8f480fd201bc6}
\BIBentrySTDinterwordspacing

\bibitem{sun_adaptive_2021}
\BIBentryALTinterwordspacing
L.~Sun, C.~Xu, Y.~He, Y.~Zhao, Y.~Xu, X.~Rui, and H.~Xu, ``Adaptive {Forest}
  {Fire} {Spread} {Simulation} {Algorithm} {Based} on {Cellular} {Automata},''
  \emph{Forests}, vol. null, p. null, 2021. [Online]. Available:
  \url{https://www.semanticscholar.org/paper/501958ef42496f1fca3b5bc07346dca2c73b6bf2}
\BIBentrySTDinterwordspacing

\bibitem{canada_forest_2013}
\BIBentryALTinterwordspacing
N.~R. Canada, ``\BIBforeignlanguage{eng}{Forest cover map},'' Jun. 2013, last
  Modified: 2023-08-21 Publisher: Natural Resources Canada. [Online].
  Available:
  \url{https://natural-resources.canada.ca/our-natural-resources/forests/sustainable-forest-management/measuring-and-reporting/remote-sensing-forestry/forest-cover-map/13433}
\BIBentrySTDinterwordspacing

\bibitem{noauthor_global_nodate}
\BIBentryALTinterwordspacing
``\BIBforeignlanguage{en-ca}{Global {Forest} {Watch} {Open} {Data} {Portal}}.''
  [Online]. Available: \url{https://data.globalforestwatch.org/}
\BIBentrySTDinterwordspacing

\bibitem{noauthor_geographic_nodate}
\BIBentryALTinterwordspacing
``Geographic {Information} {Systems} - {Forestry} {Information} {Available} for
  {Download} {\textbar} novascotia.ca.'' [Online]. Available:
  \url{https://novascotia.ca/natr/forestry/gis/downloads.asp}
\BIBentrySTDinterwordspacing

\bibitem{noauthor_canadas_nodate}
\BIBentryALTinterwordspacing
``Canada's {National} {Forest} {Information} {System}.'' [Online]. Available:
  \url{https://ca.nfis.org/index_eng.html}
\BIBentrySTDinterwordspacing

\bibitem{noauthor_topoview_nodate}
\BIBentryALTinterwordspacing
``\BIBforeignlanguage{en}{{topoView} {\textbar} {USGS}}.'' [Online]. Available:
  \url{https://ngmdb.usgs.gov/maps/topoview/}
\BIBentrySTDinterwordspacing

\bibitem{noauthor_canada_nodate}
\BIBentryALTinterwordspacing
``\BIBforeignlanguage{en}{Canada topographic map, elevation, terrain}.''
  [Online]. Available:
  \url{https://en-ca.topographic-map.com/map-kc957/Canada/}
\BIBentrySTDinterwordspacing

\bibitem{pastor2003mathematical}
E.~Pastor, L.~Z{\'a}rate, E.~Planas, and J.~Arnaldos, ``Mathematical models and
  calculation systems for the study of wildland fire behaviour,''
  \emph{Progress in Energy and Combustion Science}, vol.~29, no.~2, pp.
  139--153, 2003.

\bibitem{forthofer2007modeling}
J.~M. Forthofer, ``Modeling wind in complex terrain for use in fire spread
  prediction,'' Ph.D. dissertation, Citeseer, 2007.

\bibitem{bhakti_fire_2020}
\BIBentryALTinterwordspacing
H.~D. Bhakti, H.~Ibrahim, F.~Fristella, and M.~Faisal, ``Fire spread simulation
  using cellular automata in forest fire,'' \emph{IOP Conference Series:
  Materials Science and Engineering}, vol. 821, p. null, 2020. [Online].
  Available:
  \url{https://www.semanticscholar.org/paper/eb97ff6ccb62e5446e67b67c8e48fdd3dd5101be}
\BIBentrySTDinterwordspacing

\bibitem{shapley1953stochastic}
L.~S. Shapley, ``Stochastic games,'' \emph{Proceedings of the National Academy
  of Sciences}, vol.~39, no.~10, pp. 1095--1100, 1953.

\bibitem{littman1994markov}
M.~L. Littman, ``Markov games as a framework for multi-agent reinforcement
  learning,'' in \emph{International Conference on Machine Learning}, 1994, pp.
  157--163.

\bibitem{basar1999dynamic}
T.~Ba\c{s}ar and G.~J. Olsder, \emph{Dynamic Noncooperative Game Theory}.\hskip
  1em plus 0.5em minus 0.4em\relax SIAM, 1999, vol.~23.

\bibitem{boutilier1996planning}
C.~Boutilier, ``Planning, learning and coordination in multi-agent decision
  processes,'' in \emph{Conference on Theoretical Aspects of Rationality and
  Knowledge}, 1996, pp. 195--210.

\bibitem{wai2018multi}
H.-T. Wai, Z.~Yang, Z.~Wang, and M.~Hong, ``Multi-agent reinforcement learning
  via double averaging primal-dual optimization,'' in \emph{Advances in Neural
  Information Processing Systems}, 2018, pp. 9649--9660.

\bibitem{silver2017mastering}
D.~Silver, J.~Schrittwieser, K.~Simonyan, I.~Antonoglou, A.~Huang, A.~Guez,
  T.~Hubert, L.~Baker, M.~Lai, A.~Bolton \emph{et~al.}, ``Mastering the game of
  {Go} without human knowledge,'' \emph{Nature}, vol. 550, no. 7676, p. 354,
  2017.

\bibitem{hu2003nash}
J.~Hu and M.~P. Wellman, ``Nash {Q}-learning for general-sum stochastic
  games,'' \emph{Journal of Machine Learning Research}, vol.~4, no. Nov, pp.
  1039--1069, 2003.

\bibitem{OpenAI_dota}
OpenAI, ``Openai five,'' \url{https://blog.openai.com/openai-five/}, 2018.

\bibitem{osborne1994course}
M.~J. Osborne and A.~Rubinstein, \emph{A Course in Game Theory}.\hskip 1em plus
  0.5em minus 0.4em\relax MIT Press, 1994.

\bibitem{heinrich2015fictitious}
J.~Heinrich, M.~Lanctot, and D.~Silver, ``Fictitious self-play in
  extensive-form games,'' in \emph{International Conference on Machine
  Learning}, 2015, pp. 805--813.

\bibitem{siedler_collaborative_2022}
\BIBentryALTinterwordspacing
P.~D. Siedler, ``Collaborative {Auto}-{Curricula} {Multi}-{Agent}
  {Reinforcement} {Learning} with {Graph} {Neural} {Network} {Communication}
  {Layer} for {Open}-ended {Wildfire}-{Management} {Resource} {Distribution},''
  Apr. 2022, arXiv:2204.11350 [cs]. [Online]. Available:
  \url{http://arxiv.org/abs/2204.11350}
\BIBentrySTDinterwordspacing

\bibitem{viseras_wildfire_2021}
\BIBentryALTinterwordspacing
A.~Viseras, M.~Meissner, and J.~Marchal, ``Wildfire front monitoring with
  multiple uavs using deep q-learning,'' \emph{IEEE Access}, 2021, publisher:
  IEEE. [Online]. Available:
  \url{https://ieeexplore.ieee.org/abstract/document/9340340/}
\BIBentrySTDinterwordspacing

\bibitem{haksar_distributed_2018}
R.~N. Haksar and M.~Schwager, ``Distributed deep reinforcement learning for
  fighting forest fires with a network of aerial robots,'' in \emph{2018
  {IEEE}/{RSJ} {International} {Conference} on {Intelligent} {Robots} and
  {Systems} ({IROS})}.\hskip 1em plus 0.5em minus 0.4em\relax IEEE, 2018, pp.
  1067--1074.

\bibitem{ali_distributed_2023}
\BIBentryALTinterwordspacing
A.~Ali, R.~Ali, and M.~F. Baig, ``Distributed {Multi}-{Agent} {Deep}
  {Reinforcement} {Learning} based {Navigation} and {Control} of {UAV} {Swarm}
  for {Wildfire} {Monitoring},'' in \emph{2023 {IEEE} 4th {Annual} {Flagship}
  {India} {Council} {International} {Subsections} {Conference}
  ({INDISCON})}.\hskip 1em plus 0.5em minus 0.4em\relax IEEE, 2023, pp. 1--8.
  [Online]. Available:
  \url{https://ieeexplore.ieee.org/abstract/document/10270198/}
\BIBentrySTDinterwordspacing

\end{thebibliography}

\end{document}